\documentclass[twocolumn]{svjour3}
\usepackage[utf8]{inputenc}
\usepackage[T1]{fontenc}
\usepackage{graphicx}
\usepackage{grffile}
\usepackage{longtable}
\usepackage{wrapfig}
\usepackage{rotating}
\usepackage[normalem]{ulem}
\usepackage{amsmath}
\usepackage{textcomp}
\usepackage{amssymb}
\usepackage{capt-of}
\usepackage{hyperref}
\usepackage{tikz}
\usepackage{subcaption}
\usepackage{subcaption}
\usepackage{placeins}
\usepackage[combine]{ucs}
\usepackage{amsmath}
\usepackage{mathtools}
\usepackage{grffile}
\usepackage{float}
\usepackage{graphicx}
\usepackage{bm}
\usepackage[numbers]{natbib}
\journalname{Progress in Artificial Intelligence}
\makeatletter
\renewcommand\tagform@[1]{\maketag@@@{\ignorespaces#1\unskip\@@italiccorr}}
\makeatother
\newcommand{\norm}[1]{\left\lVert#1\right\rVert}
\newcommand{\vect}[1]{\boldsymbol{#1}}
\DeclareMathOperator*{\argmax}{arg\,max}

\begin{document}
\institute{Bjørn Magnus Mathisen \at EXPOSED Aquaculture Research Centre \\ Department of Computer Science \\ Trondheim, Norway \\ \url{http://www.idi.ntnu.no} \\ \email{\{bjornmm,agnar,kerstin.bach,helgel\}@ntnu.no}}
\authorrunning{Mathisen}
\author{Bjørn Magnus Mathisen \href{https://orcid.org/0000-0002-8063-1835}{\includegraphics[scale=0.7]{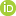}}, Agnar Aamodt\href{https://orcid.org/0000-0003-0114-8931}{\includegraphics[scale=0.7]{Fig0-orcid.png}}, Kerstin Bach \href{https://orcid.org/0000-0002-4256-7676}{\includegraphics[scale=0.7]{Fig0-orcid.png}} and Helge Langseth \href{https://orcid.org/0000-0001-6324-6284}{\includegraphics[scale=0.7]{Fig0-orcid.png}}}
\date{\today}
\title{Learning similarity measures from data}
\hypersetup{
 pdfauthor={Bjørn Magnus Mathisen \href{https://orcid.org/0000-0002-8063-1835}{\includegraphics[scale=0.7]{Fig0-orcid.png}}, Agnar Aamodt\href{https://orcid.org/0000-0003-0114-8931}{\includegraphics[scale=0.7]{Fig0-orcid.png}}, Kerstin Bach \href{https://orcid.org/0000-0002-4256-7676}{\includegraphics[scale=0.7]{Fig0-orcid.png}} and Helge Langseth \href{https://orcid.org/0000-0001-6324-6284}{\includegraphics[scale=0.7]{Fig0-orcid.png}}},
 pdftitle={Learning similarity measures from data},
 pdfkeywords={},
 pdfsubject={},
 pdfcreator={Emacs 28.0.50 (Org mode 9.3.1)}, 
 pdflang={English}}

\maketitle
\begin{abstract}
Defining similarity measures is a requirement for some machine learning methods.
One such method is case-based reasoning (CBR) where the similarity measure is
used to retrieve the stored case or set of cases most similar to the query case.
Describing a similarity measure analytically is challenging, even for domain
experts working with CBR experts. However, data sets are typically gathered as
part of constructing a CBR or machine learning system. These datasets are
assumed to contain the features that correctly identify the solution from the
problem features, thus they may also contain the knowledge to construct or learn
such a similarity measure. The main motivation for this work is to automate the
construction of similarity measures using machine learning. Additionally, we
would like to do this while keeping training time as low as possible. Working
towards this, our objective is to investigate how to apply machine learning to
effectively learn a similarity measure. Such a learned similarity measure could
be used for CBR systems, but also for clustering data in semi-supervised
learning, or one-shot learning tasks. Recent work has advanced towards this
goal, relies on either very long training times or manually modeling parts of
the similarity measure. We created a framework to help us analyze current
methods for learning similarity measures. This analysis resulted in two novel
similarity measure designs. One design using a pre-trained classifier as basis
for a similarity measure. The second design uses as little modeling as possible
while learning the similarity measure from data and keeping training time low.
Both similarity measures were evaluated on 14 different datasets. The evaluation
shows that using a classifier as basis for a similarity measure gives state of
the art performance. Finally the evaluation shows that our fully data-driven
similarity measure design outperforms state of the art methods while keeping
training time low. \noindent
\end{abstract}

\thanks{This work was supported by the Research Council of Norway through the
EXPOSED project(grant number 302002390) and the Norwegian Open AI Lab}

\section{Introduction}
\label{sec:orgb6d392f}
\label{sec:introduction}
Many artificial intelligence and machine learning (ML) methods, such as
k-nearest neighbors (k-NN) rely on a similarity (or distance) measure
\cite{maggini2012learning} between data points. In Case-based reasoning (CBR) a
simple k-NN or a more complex similarity function is used to retrieve the stored
cases that are most similar to the current query case. The similarity measure
used in CBR systems for this purpose is typically built as a weighted Euclidean
similarity measure (or as a weight matrix for discrete and symbolic variables).
Such a similarity measure is designed with assistance of domain experts by
adjusting the weights for each attribute of the cases to represent how important
they are (one example can be seen in \cite{Wienhofen2016}, or generally
described in chapter 4 of \cite{bergmann2002experience})

In many situations the design of such a function is non-trivial. Domain experts
with an understanding of CBR or machine learning are not easily available.
However, before or during most CBR projects, data is gathered that relates to
the problem being solved by the CBR system. This data is used to construct cases
for populating the case base. If the data is labeled according to the
solution/class, it can be used to learn a similarity measure that is relevant to
the task being solved by the system. Exploring efficient methods of learning
similarity measures and improving on them is the main motivation of this work.

\begin{figure}[!ht]
\centering
\includegraphics[width=0.8\linewidth]{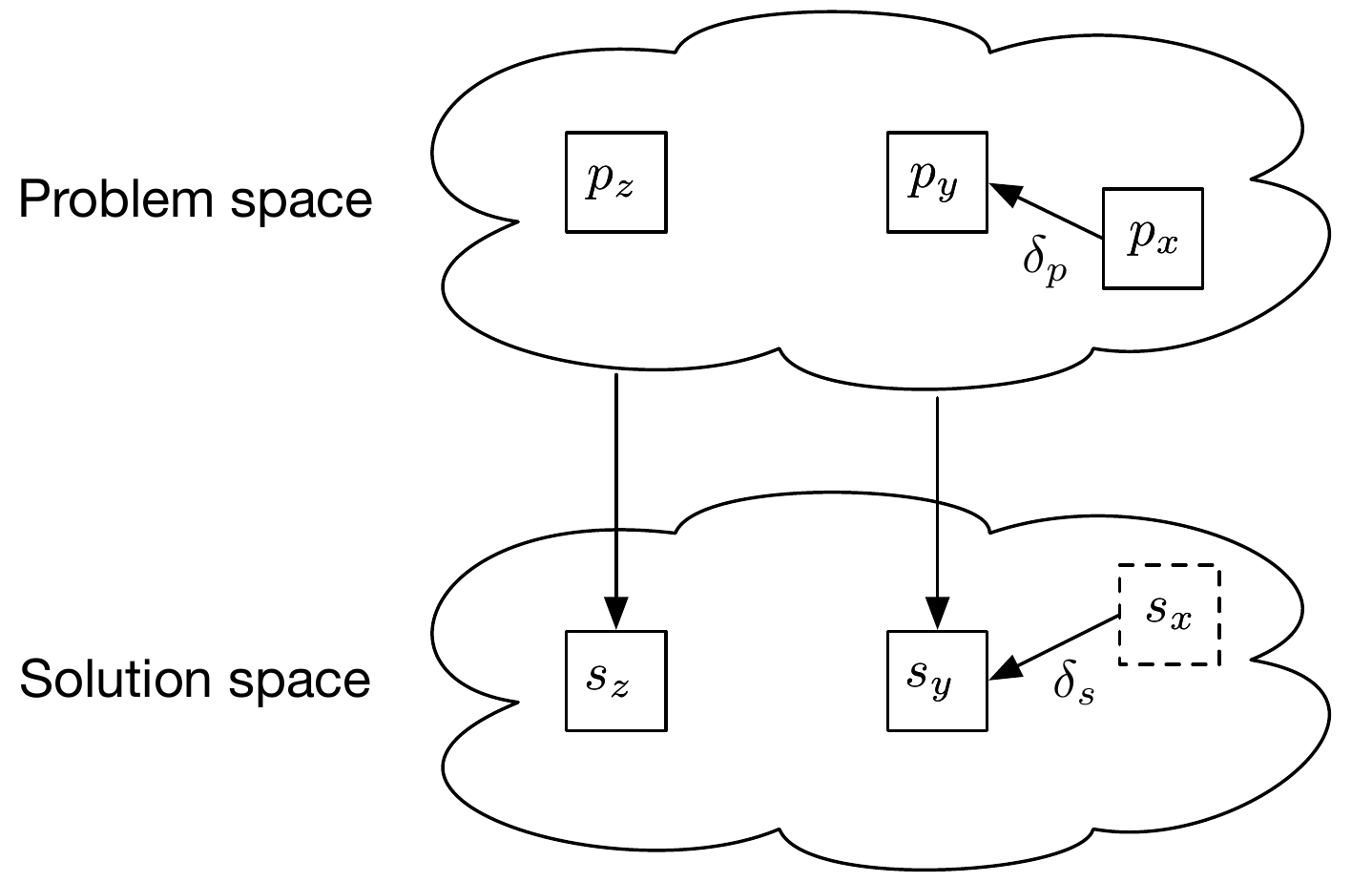}
\caption[leake1996case]{\label{fig:problem-solution-space}Illustration of problem and solution spaces \cite{leake1996case}. \(p_{y}\) and \(p_{z}\) are two problem descriptions with features describing a problem each of which has a corresponding (\(s_{y}\) and \(s_{z}\)) solution in solution space. \(\delta_{p}\) illustrates the distance between a new problem \(p_{x}\) and a stored problem \(p_{y}\). Correspondingly \(\delta_{s}\) is the distance between the solution \(s_{y}\) and the solution \(s_{x}\) which is the (unknown) ideal solution to \(p_{x}\). A fundamental assumption in CBR is that if the similarity between \(p_{x}\) and \(p_{y}\) is high then the similarity between the unknown solution \(s_{x}\) to \(p_{y}\) is high (\(\delta_{p} \approx \delta_{s}\)): Similar problems have similar solutions.}
\end{figure}

In the CBR literature, similarity measurement is often described in terms of
problem- and solution spaces. Problem space is where the features of a problem
describe the problem; this is often called feature space in non-CBR ML
literature. Solution space, also referred to as target space, is populated by
points describing solutions to points in the problem space. The function that
maps a point from the problem space to its corresponding point in the solution
space is typically the goal of supervised machine learning. This is illustrated
in Figure \ref{fig:problem-solution-space}.

A similarity measure in the problem space represents an approximation of the
similarity between two cases or data points in the solution space (i.e. whether
these two cases have similar or dissimilar solutions). Such a similarity measure
would be of great help in situations where lots of labeled data is available,
but domain knowledge is not available, or when the modeling of such a similarity
measure is too complex.

Learned similarity measures can also be used in other settings, such as
clustering. Another relevant method type is semi-supervised learning in which
the labeled part of a dataset is used to cluster or label the unlabeled part.

How to automatically learn similarity measures has been an active area of
research in CBR. For instance, Gabel et al. \cite{gabel2015ann} train a similarity
measure by creating a dataset of collated pairs of data points and their
respective similarities. This dataset is then used to train a neural network to
represent the similarity measure. In this method the network needs to extract
the most important features in terms of similarity for both data points, then
combine these features to output a similarity measure. Recent work (e.g. Martin
et al. \cite{martin2017convolutional}) has used Siamese neural networks (SNN)
\cite{bromley1994signature} to learn a similarity measure in CBR. SNNs have the
advantage of sharing weights between two parts of the network, in this case the
two parts that extract the useful information from the two data points being
compared. All of these methods for learning similarity measures have in common
that they are trained to compare two data points and return a similarity
measurement. Our work of automatically learning similarity measures is also
related to the work done by H{\"u}llermeier et al. on preference-based
CBR \cite{hullermeier2011preference,hullermeier2013preference}. In this work the
authors learn a preference of similarity between cases/data points, which
represents a more continuous space between solutions than a typical similarity
measure in CBR. This type of approach to similarity measures is similar to
learning similarity measures by using machine learning models, in that both can
always return a list of possible solutions sorted by their similarity.

In this work we have developed a framework to show the main differences between
various types of similarity measures. Using this framework, we highlight the
differences between existing approaches in Section \ref{sec:relatedwork}. This
analysis also reveals areas that have not received much attention in the
research community so far. Based on this we developed two novel designs for
using machine learning to learn similarity measures from data. Both of the two
designs are continuous in their representation of the estimated solution space.

The novelty of our work is three-fold: First showing that using a classifier as
a basis for a similarity measure gives adequate performance. Then we demonstrate
similarity measure designed to use as little modeling as possible, while keeping
training time low, outperforms state of the art methods. Finally to analyze the
state of the art and compare it to our new similarity measure design we
introduce a simple mathematical framework. We show how this is a useful tool for
analyzing and categorizing similarity measures.

The remainder of this paper describes our method in more detail. Section
\ref{sec:framework} describes the novel framework for similarity measurement
learning, and Section \ref{sec:relatedwork} then summarizes previous relevant work
in relation to this framework. In Section \ref{sec:method} we describe
suggestions of new similarity measures, and how we design the experimental
evaluation. Subsequently, in Section \ref{sec:evaluation} we show the results of
this evaluation. Finally, in Section \ref{sec:conclusion} we interpret and discuss
the evaluation results and give some conclusions. We present some of the
limitations of our work as well as possible future paths of research.

\section{A framework for similarity measures}
\label{sec:org409539d}
\label{sec:framework}

\begin{figure}[!ht]
\centering
\includegraphics[width=0.8\linewidth]{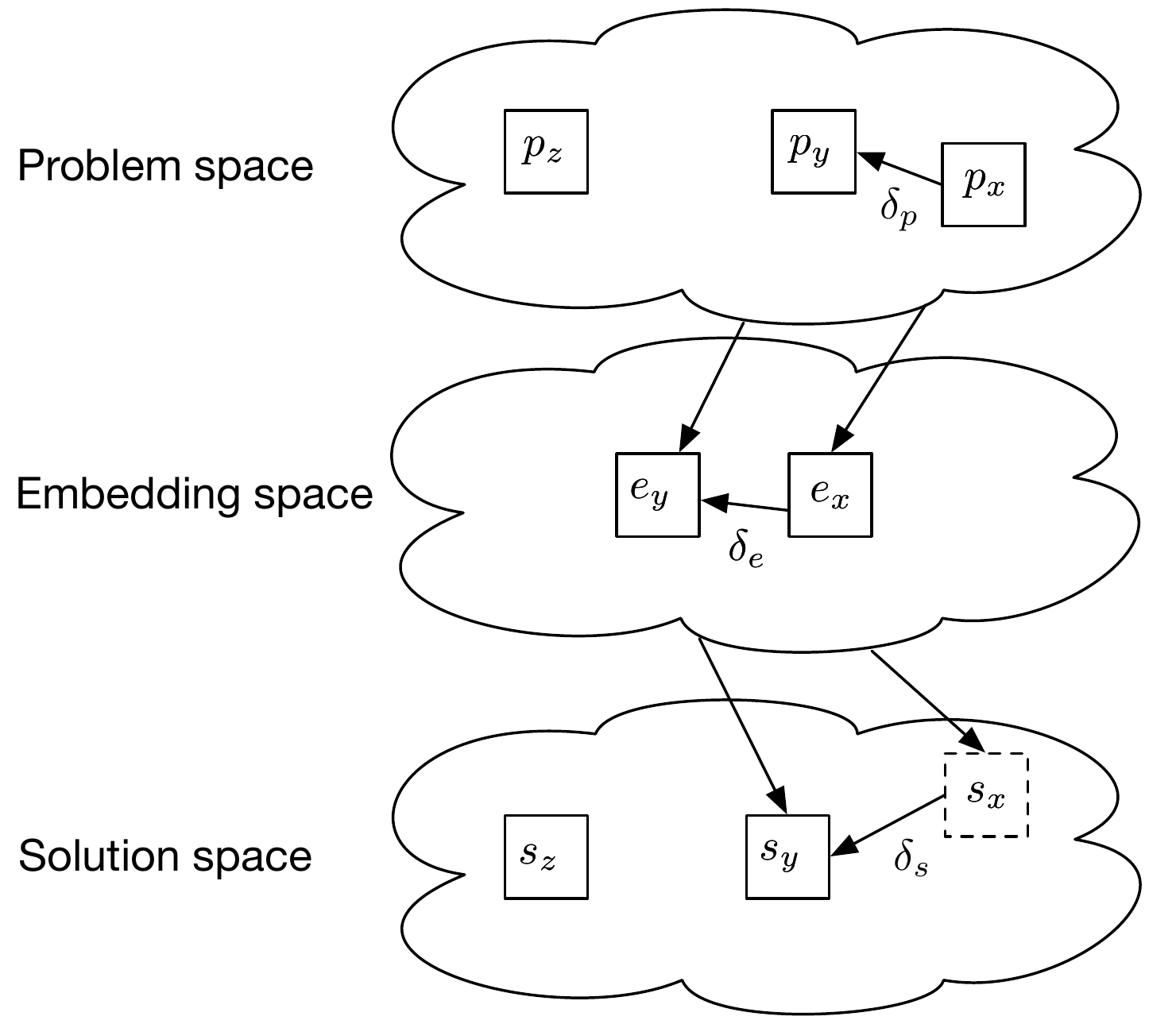}
\caption[eq:bmmsim leake1996case]{\label{fig:problem-embedding-solution-space}Illustrating how \(G(\cdot)\) from Equation \ref{eq:bmmsim} adds another space, the embedding space, between the problem and the solution space \cite{leake1996case} (see Figure \ref{fig:problem-solution-space}). \(C(\cdot)\) then combines the two embeddings of \(p_{y}\) and \(p_{x}\) (\(e_{y}\) and \(e_{x}\) respectively) and calculates the similarity \(\delta_{e}\) between them. The main assumption is that distance in embedding space (\(\delta_{e}\)) is close to the distance in solution space (\(\delta_{s}\)) ; if the embedded points \(e_{x}\) and \(e_{y}\) are similar, then the true (unknown) solution \(s_{x}\) is similar to solution \(s_{y}\). The main contribution of \(G(\cdot)\) is to create a embedding space such that the distance in embeddings space (\(\delta_{e}\)) is a better estimate of the distance in solution space (\(\delta_{s}\)) than the distance in problem space (\(\delta_{p}\)).}
\end{figure}

We suggest a framework for analyzing different functions for similarity with
\(\mathbb{S}\) as a similarity measure applied to pairs of data points
\((\vect{x},\vect{y})\); 

\begin{equation}
\label{eq:bmmsim}
\mathbb{S}(\vect{x},\vect{y}) = C(G(\vect{x}),G(\vect{y})) ,
\end{equation}

\noindent where \(G(\vect{x}) = \hat{\vect{x}}\) and \(G(\vect{y}) =
\hat{\vect{y}}\) represents embedding or information extraction from data points
\(x\) and \(y\) , i.e. \(G(\cdot)\) highlights the parts of the data points most
useful to calculate the similarity between them. An illustration of this process
can be seen in Figure \ref{fig:problem-embedding-solution-space}.

\(C(G(\vect{x}),G(\vect{y})) = C(\hat{\vect{x}},\hat{\vect{y}})\) models the
distance between the two data points based on the embeddings \(\hat{\vect{x}}\)
and \(\hat{\vect{y}}\). The functions \(C\) and \(G\) can be either manually modeled
or learned from data. With respect to this we will enumerate all of the
different configurations of Equation \ref{eq:bmmsim} and describe their main
properties and how they have been implemented in state of the art research. Note
that we will use \(\mathbb{S}(\cdot)\) to annotate the similarity measurement and
\(C(\cdot)\) for the sub-part of the similarity measurement that calculates the
distance between the two outputs of \(G(\cdot)\). \(\mathbb{S}(\cdot)\) is distinct
from \(C(\cdot)\) unless \(G(x) = I(x) = x\).

\begin{table}[htbp]
\begin{center}
\begin{tabular}{|l|l|l|l|}
\hline
\multicolumn{2}{|l|}{} & \multicolumn{2}{l|}{$C(\hat{\vect{x}},\hat{\vect{y}})$} \\
\cline{3-4}
\multicolumn{2}{|l|}{} & Modeled & Learned \\
\hline
$G(\cdot)$ & Modeled & Type 1 & Type 2 \\
\cline{2-4}
 & Learned & Type 3 & Type 4 \\
\hline
\end{tabular}
\caption{\label{table:typematrix}
Table showing different types of similarity measures in our proposed framework.}
\end{center}
\end{table}

In the following we characterize the different types of similarity measures:

\begin{description}
\item[{\textbf{Type 1}}] A typical similarity measure in CBR systems would model
\(C(\hat{\vect{x}},\hat{\vect{y}})\) and \(G(\cdot)\) from domain
knowledge. Such a similarity measure is typically modeled by
experts with the relevant domain knowledge together with CBR
experts, who know how to encode this domain knowledge into
the similarity measures.

For example when modeling the similarity measure of cars for
sale, where the goal is to model the similarity of cars in
terms of their final selling price. In this example, domain
experts may model the embedding function \(G(\cdot)\) so that
the amount of miles driven has a greater importance than the
color of the car. \(C(\hat{\vect{x}},\hat{\vect{y}})\) could be
modeled such that difference in miles driven is less
important than difference the number of repairs done on the
car. More details and examples can be found in
\cite{cunningham2009taxonomy}.
\item[{\textbf{Type 2}}] This type represents similarity measures that models
\(G(\cdot)\) and learns the function
\(C(\hat{\vect{x}},\hat{\vect{y}})\). In this context
\(G(\cdot)\) can be viewed as an embedding function. Since
\(G(\cdot)\) is not learned from the data it is not interesting
to analyze it as part of learning the similarity measure, as
processing the data through \(G(\cdot)\) could be done in batch
before applying the data to \(\mathbb{S}(\vect{x},\vect{y})\).
Learning \(C(\hat{\vect{x}},\hat{\vect{y}})\) needs to be done
with a dataset consisting of triplets of the data points
\(\hat{\vect{x}}\) and \(\hat{\vect{y}}\), and \(s\) being the true
similarity between \(\hat{\vect{x}}\) and \(\hat{\vect{y}}\).

A special case of Type 2 is when \(G(\cdot)\) is set to be the
identity function \(I(\vect{x})=G(\vect{x})=\vect{x}\), while
\(C(\vect{x},\vect{y})\) is learned from data. Examples of this
type are presented for example in Gabel et al.
\cite{gabel2015ann} where the similarity measure always looks
at the two inputs together, never separately.

\item[{\textbf{Type 3}}] In this type of similarity measure the embedding/feature
extraction \(G(\cdot)\) is learned and
\(C(\hat{\vect{x}},\hat{\vect{y}})\) is modeled. Typically the
embedding function learned by \(G(\cdot)\) resembles the
function that is the goal during supervised machine learning.
Within the similarity measurement \(\hat{\vect{x}} = G(x)\) is
used as an embedding vector for calculating similarity, when
in classification \(\hat{\vect{x}}\) would be the softmax
vector output. Using a pre-trained classification model as a
starting point for \(G(\vect{x}) = \hat{\vect{x}}\) as input to
e.g. \(C(\hat{\vect{x}},\hat{\vect{y}}) =
                   \norm{\hat{\vect{x}} - \hat{\vect{y}}}_{1}\) should give good
results for similarity measurements if that model had high
precision for classification within the same dataset.

However it is not given that the best embedding vector for
calculating similarity is the same as the embedding vector
produced by a \(G(x)\) trained to do classification.

\item[{\textbf{Type 4}}] This measure is designed so that both \(G(\cdot)\) and
\(C(\hat{\vect{x}},\hat{\vect{y}})\) are learned.
\end{description}

We will design, implement and evaluate similarity measures based on Type 1,
Type 3, Type 2 and Type 4 in Section \ref{sec:method}. These results
will be shown in Sections \ref{sec:evaluation}.

To allow \(\mathbb{S}\) as a similarity measurement for clustering e.g. k-nearest
neighbors, a similarity measure should fulfill the following requirements:

\begin{description}
\item[{\textbf{Symmetry}}] \label{symmetry}\(\mathbb{S}(\vect{x},\vect{y}) = \mathbb{S}(\vect{y},\vect{x})\)
The similarity between \(\vect{x}\) and \(\vect{y}\) should be the
same as the similarity between \(\vect{x}\) and \(\vect{y}\).
\item[{\textbf{Non-negative}}] \(\mathbb{S}(\vect{x},\vect{y}) \geq 0 | \forall \vect{x},\vect{y}\) The
similarity between to data-points can not be negative.
\item[{\textbf{Identity}}] \(\mathbb{S}(\vect{x},\vect{y}) = 1 \Longleftrightarrow \vect{x}
                = \vect{y}\) The similarity between two data-points should be 1
iff \(\vect{x}\) is equal to \(\vect{y}\).
\end{description}

Some of these requirements are not satisfied by all types of similarity
measures, i.e. symmetry is not a direct design consequence of Type 2 but of
Type 3 if \(C(\hat{\vect{x}},\hat{\vect{y}})\) is symmetric. Even if symmetry
is not present in all similarity measures \cite{tversky1977features} it is
important for reducing training time, as the training set size goes from
\(N(N-1)\) to \(N(\frac{N}{2} - 1)\). Symmetry also enables the similarity measure to
be used for clustering.

In the next section, we will relate current state of the art to the framework in
context of the different types.

\section{Related work}
\label{sec:orgb81a0c6}
\label{sec:relatedwork}

To exemplify the framework presented in the previous section we will relate
previous work to the framework and the types of similarity measurements that
derive from the framework. This will also enable us to see possibilities for
improvement and further research. 

As stated in Section \ref{sec:introduction} our motivation is to automate the
construction of similarity measures. Additionally, we would like to do this
while keeping training time as low as possible. Thus we will not focus on
Type 1 similarity measures as this type uses no learning. Both Type 2
and Type 4 require a different type of training dataset than a typical
supervised machine learning dataset, as \(C(\vect{x},\vect{y})\) is typically
dependent on the order of the data points (see Section \ref{sec:method}). Thus
given our initial motivation, Type 3 similarity measures seems to be the
most promising type of similarity measure to focus on. However, it is worth
investigating similarity measures of Type 4, to see if the added benefit of
learning \(C(\vect{x},\vect{y})\) outweighs the added training time. Or if it is
possible to make it symmetric (as defined in the previous section) so that the
training time could become similar to Type 3.

Thus we will focus on summarizing related work from Type 3 similarity
measures, but also add relevant work from Type 1, Type 2 and
Type 4 for reference.

Type 1 is a type of similarity measure which is manually constructed. A
general overview and examples of this type of similarity measure can be found in
\cite{cunningham2009taxonomy}. Nikpour et al. \cite{nikpour2018bayesian}
presents an alternative method which includes enrichment of the cases/data points
via Bayesian networks.

\textbf{Type 2}

In Type 2 similarity measures only the binary \(C(\vect{x},\vect{y})\)
operator of the similarity measure \(\mathbb{S}(\vect{x},\vect{y})\) is learned,
while \(G(\cdot)\) is either modeled or left as the identity function
(\(G(\vect{x}) = I(\vect{x}) = \vect{x}\)). Stahl et al. have done a lot of work on
learning Type 2 similarity measures from data or user feedback. In all of
their work they formulate \(C(\vect{x},\vect{y}) = \sum \vect{w}_{i} *
sim_{i}(\vect{x}_{i},\vect{y}_{i})\) where for each feature \(i\), \(sim_{i}\) is the
local similarity measure and \(\vect{w}_{i}\) is the weight of that feature. In
\cite{stahl2001learning} Stahl et al. describe a method for learning the feature
weights.

In \cite{stahl2003using} Stahl et al. introduce learning local similarity measures
through an evolutionary algorithm (EA). First they learn attribute weights
(\(\vect{w}_{i}\)) by adopting the method previously described in
\cite{stahl2001learning}. Then they use an EA to learn the local similarity
measures for each feature (\(sim_{i}(\vect{x},\vect{y})\)). In
\cite{stahl2006optimizing} Stahl and Gabel present work were they learn weights of
a modeled similarity measure, and the local similarity for each attribute
through an ANN. Reategui et al. \cite{reategui1997combining} learn and represent
parts of the similarity functions (\(C(\hat{\vect{x}},\hat{\vect{y}})\)) through
ANN. Langseth et al. \cite{langseth1999learning} learn similarity knowledge
(\(C(\hat{\vect{x}},\hat{\vect{y}})\)) from data using Bayesian networks, which
still partially relies on modeling the Bayesian networks with domain knowledge.

Abdel-Aziz et al. \cite{abdel2014learning} use the distribution of case attribute
values to inform a polynomial local similarity function, which is better than
guessing when domain knowledge is missing. So this method extracts statistical
properties from the dataset to parametrize \(C(\hat{\vect{x}},\hat{\vect{y}})\).

Gabel and Godehardt \cite{gabel2015ann} use a neural network to learn a similarity
measure. Their work is done in the context of Case-based Reasoning (CBR) which
uses the measure to retrieve similar cases. They concatenate the two data points
into one input vector. Thus in the context of our framework \(G(\cdot)\) is
modeled as a identify function \(I(x) = x\) and \(C(\vect{x},\vect{y})\) is
learned.

Maggini et al. \cite{maggini2012learning} uses SIMNNs which they also see as a
special case of the Symmetry Networks \cite{shawe1993symmetries} (SNs). In SIMMNs
\(C(\hat{\vect{x}},\hat{\vect{y}})\) and \(G(\cdot)\) are both a function of both
\(\vect{x}\) and \(\vect{y}\) data points and there is thus no distinct \(G(\cdot)\).
They also have a specialized structure imposed on their network to make sure the
learned function is symmetric. SIMNN is in essence an extended version of a
Siamese neural network, but without a distinct distance layer usually present in
SNN architectures. They focus on the specific properties of the network
architecture and the application of such networks in semi-supervised settings
such as k-means clustering. The pair of data points (\(\vect{x}\) and \(\vect{y}\))
are being compared two times, the first time at the first hidden layer, then at
the output layer. Since there are no learnable parameters before this comparison
all the learning is done in \(C(\hat{\vect{x}},\hat{\vect{y}})\) and \(G(\vect{x})\)
is the activation function of the input layer.

\textbf{Type 3}

One way of looking at a similarity measure is as an inverse distance measure, as
similarity is the semantic opposite of distance. There has been much work on
learning distance measures. Most of this work can be categorized as a
Type 3 similarity measure as the learning tasks only aims to learn the
embedding function \(G(\cdot)\) then combine the output of this function with a
static \(C(\cdot)\) (e.g. a \(L2\) norm function). The most well known instance of a
Type 3 learned distance measure is Siamese neural networks (SNNs), it is
highly related to the Type 2 similarity measure by Maggini et al.'s
Similarity neural networks (SIMNN) \cite{maggini2012learning}.

The main characteristic of SNNs is sharing the weights between the two identical
neural networks. The data points we want to measure the similarity for are then
input to these networks. This frees the learning algorithm of learning two sets
of weights for the same task. This was first used in \cite{bromley1994signature}
(using \(C(\hat{\vect{x}},\hat{\vect{y}}) = cos(\hat{\vect{x}},\hat{\vect{y}})\)
and \(G(\cdot)\) being learned from data) to measure similarity between
signatures. Similar architectures are also discussed in
\cite{shawe1993symmetries}.

Chopra et al. \cite{chopra2005learning} uses a SNN for face verification, and pose
the problem as an energy based model. The output of the SNN are combined through
a \(L1\) norm (absolute-value norm \(C(\hat{\vect{x}},\hat{\vect{y}}) =
\norm{\hat{\vect{x}} - \hat{\vect{y}}}\)) to calculate the similarity. They
emphasize that using a \(L2\) norm (Euclidean distance) as part of the loss function
would make the gradient too small for effective gradient descent (i.e. create
plateaus in the loss function). This work is closely related to Hadsell et al.
\cite{hadsell2006dimensionality}, where they explain the contrastive loss function
used for training the SNN (also used in
\cite{chopra2005learning,martin2017convolutional}) by analogy of a spring system.

Related to this Vinyals et al. \cite{vinyals2016matching} uses a similar type of
setup for matching an input data point to a support set. It is framed as a
discriminative task, where they use two neural networks to parametrize an
attention mechanism. They use these two networks to embed the two data points
into a feature space where the similarity between them are measured. However, in
contrast to SNNs and SIMNNs, their two networks for embedding the data points
are not identical, as one network is tailored to embed a data point from the
support set, but also given the rest of the support set. Thus the embedding of
the support set data point is also a function of the rest of the support set.
With \(C(\hat{\vect{x}},\hat{\vect{y}})\) being modeled as a cosine softmax, this
is similar to the examples of Type 3 similarity measures mentioned
previously (e.g. \cite{bromley1994signature,berlemont2015siamese}). However a
major difference is that signal extraction functions are not equal:
\(\mathbb{S}(\vect{x},\vect{y}) = C(f(\vect{x}),g(\vect{x}))\) with \(f(\vect{x})
\neq g(\vect{x})\) (only stating that \(f(\cdot)\) may potentially equal
\(g(\cdot)\)). Since \(f(\cdot)\) and \(g(\cdot)\) are not sharing weights between
them, the architecture is variant (or asymmetric) to the ordering of input
pairs. Thus the architecture needs up to twice as much training to achieve the
same performance as a SNN.

In much of the same fashion as Chopra et al. did in \cite{chopra2005learning},
Berlemot et al. \cite{berlemont2015siamese} uses SNNs combined with an energy based
model to build a similarity measure between different gestures made with
smart phones. However they adapt the error estimation from using only separate
positive and negative pairs to a training subset including; a reference sample,
a positive sample and a negative sample for every other class. They train
\(G(\cdot)\) while keeping a static \(C(\hat{\vect{x}},\hat{\vect{y}}) =
cos(\hat{\vect{x}},\hat{\vect{y}})\). This training method of using triplets for
training SNNs was also described by Lefebvre et al. \cite{lefebvre2013learning}. A
similar approach can be seen in Hoffer et al. \cite{hoffer2015deep}, however they
do not use a set of negative examples per reference point for each class as
Berlemont et al do. Instead they use triples of
\((\vect{x},\vect{x}^{+},\vect{x}^{-})\), \(\vect{x}\) being the reference point,
\(\vect{x}^{+}\) being the same class and \(\vect{x}^{-}\) being a different class.

Koch et al. \cite{koch2015siamese} uses a Convolutional Siamese Network (CSN),
with \(G(\cdot)\) implemented as a CNN and \(C(\hat{\vect{x}},\hat{\vect{y}})\)
implemented as \(L1(\hat{\vect{x}},\hat{\vect{y}})\). This is done in a
semi-supervised fashion for one-shot learning within image recognition. They
learn this CSN for determining if two pictures from the Omniglot
\cite{lake2015human} dataset is within the same class. The model can then be used
to classify a data point representing an unseen class by comparing it to a
repository of class representatives (Support Set).

CSNs are also used in the context of CBR by Martin et al.
\cite{martin2017convolutional} to represent a similarity measure in a CBR system.
The CSN is trained with pairs of cases and the output is their similarity.
During training they have to label pairs of cases as 'genuine' (both cases
belong to the same class) or 'impostor' (the cases belong to different classes).
This requires that the user has a clear boundary for the classes. In relation to
our framework this similarity measure learns \(G(\cdot)\), while
\(C(\hat{\vect{x}},\hat{\vect{y}})\) is static. With \(G(\cdot)\) implemented as a
convolutional neural network, and \(C(\hat{\vect{x}},\hat{\vect{y}})\) implemented
as Euclidean distance (\(L2\) norm).

In general using SNNs for constructing similarity measures have a major
advantage as you can easily adopt pre-trained models for \(G(\cdot)\) to
embedding/pre-process the data points. For example to train a model for
comparing two images one could use ResNet \cite{he2016deep} for \(G(\cdot)\) then
use the \(L1\) norm as \(C(\hat{\vect{x}},\hat{\vect{y}})\). This would be a very
similar approach to the similarity measure used by Koch et al.
\cite{koch2015siamese} with \(\mathbb{S}(\vect{x},\vect{y}) =
\norm{(G(\vect{x}),G(\vect{y}))}_{1}\), the main difference being that \(G(\cdot)\)
is designed for bigger pictures.

There are only very few examples of Type 4 similarity measures in the
literature. In Zagoruyko and Komodakis's work \cite{zagoruyko2015learning} they
investigate different types of architectures for learning image comparison using
convolutional neural networks. In all of the architectures they evaluate
\(C(\hat{\vect{x}},\hat{\vect{y}})\) is learned, but in some of these
architectures \(G(\cdot)\) is not symmetric, i.e. \(\mathbb{S}(\vect{x},\vect{y}) =
C(G(\vect{x}),H(\vect{y}))\) where \(G(\vect{x}) \neq H(\vect{x})\).
Arandjelovi\'{c} and Zisserman's work \cite{arandjelovic2017look} use a
very similar method to many Type 3 similarity measures for calculating
similarity. However their input data is always pairs of two different data types
and is as such different from most of the other relevant work leaving \(G(\cdot)\)
unsymmetrical just as in Zgoruyko et al. \cite{zagoruyko2015learning} and Vinyals
et al. \cite{vinyals2016matching}. In contrast to the Type 3 similarity
measures including \cite{vinyals2016matching}, Arandjelovi\'{c} et al.
also learns \(C(\hat{\vect{x}},\hat{\vect{y}})\), which they call a fusion layer.

All similarity measure of Type 3 we found in the literature use a loss
function that includes feedback from the binary operator part of \(\mathbb{S}\)
(\(C(\hat{\vect{x}},\hat{\vect{y}})\)). In the case of SNNs even if
\(C(\vect{x},\vect{y})\) is non-symmetric (\(C(\vect{x},\vect{y}) \neq
C(\vect{y},\vect{x})\)) the loss for each network would be equal as they are
equal and share weights. That means that ordering of the two data points being
compared during training has no effect, i.e. the training effect of
\((\vect{x},\vect{y})\) is equal to that of \((\vect{y},\vect{x})\). This means a
lot of saved time during training, as the training dataset could be halved
without any negative effect on performance.

However the implementation of \(C(\hat{\vect{x}},\hat{\vect{y}})\) would then
decide how much training one would need to adapt a pre-trained model from
classifying single data points to measuring similarity between them. One could
view the process of starting with a pre trained model for the dataset, then
training the model with loss coming from \(C(\hat{\vect{x}},\hat{\vect{y}})\) as
adapting the model from classification to similarity measurement.

One way of creating a Type 3 similarity measure using a minimal amount of
training would be to pre-train a network on classifying individual data points.
Then apply that network as \(G(\cdot)\) that feeds into a
\(C(\hat{\vect{x}},\hat{\vect{y}}) = \norm{\hat{\vect{x}} - \hat{\vect{y}}}\) in a
similarity measurement. Evaluation of such a similarity measurement has not been
reported in literature, and such a similarity will be explored in the next
section.

\section{Method}
\label{sec:org84877f0}
\label{sec:method}

The framework presented in Section \ref{sec:framework} and the subsequent analysis
of previous relevant work presented in Section \ref{sec:relatedwork} shows that
there are unexplored opportunities within research on similarity measurements.

Given the initial motivation we seek methods that work well in domains where
domain knowledge is very resource demanding. This requires that as much as
possible of the similarity measure \(\mathbb{S}(\vect{x},\vect{y}) =
C(G(\hat{\vect{x}}),G(\hat{\vect{y}}))\) is learned from data rather than modeled from domain
knowledge. There are some exceptions to this, such as applying general binary
operations, such as norms (e.g. \(L1\) or \(L2\) norm), on the two data points
(\(\hat{\vect{x}}\) and \(\hat{\vect{y}}\)) pre-processed by \(G(\cdot)\). In these
cases there is little domain expertise involved in designing
\(C(\hat{\vect{x}},\hat{\vect{y}})\) other than intuition that the similarity of
two data points is closely related to the norm between \(\hat{\vect{x}}\) and
\(\hat{\vect{y}}\).

The most promising type of similarity measures from this perspective are
Type 3 and Type 4 where \(G(\cdot)\) is learned in Type 3 and both
\(C(\vect{x},\vect{y})\) and \(G(\cdot)\) are learned in Type 4. However, to
test any new design we need to have reference methods to compare against. For
reference, we chose to implement one Type 1 similarity measure, two
similarity measures of Type 2 (including Gabel et. al's) similarity
measures and Chopra et. al's Type 3 similarity measure. The Type 1
similarity measure uses a similarity measure that weights each feature
uniformly. The Type 2 is identical to the Type 1 similarity measure
except that it uses a local similarity function for each feature which is
parametrized by statistical properties of the values of that feature in the
dataset.

One unexplored direction of creating similarity measures is creating a SNN
similarity measure (Type 3) through training \(G(\cdot)\) as a classifier on
the dataset later being used for measuring similarity. Then using that trained
\(G(\cdot)\) to construct a SNN similarity measure. This is in contrast to the
usual way of training SNNs (as seen in e.g
\cite{chopra2005learning,bromley1994signature}) where the loss function is a
function of pairs of data points, not single data points. The motivation for
exploring this type of design is that it shows the similarity measuring performance
of using networks pre-trained on classifying data points directly as part of a
SNN similarity measure. This will be detailed in Subsection \ref{sec:type3impl}.

Finally, we will explore Type 4 similarity measures which have seen little
attention in research so far. To make our design as symmetric as possible we
will use the same design as SNNs for \(G(\cdot)\) and introduce novel design to
also make \(C(\hat{\vect{x}},\hat{\vect{y}})\) symmetric. That way our design is
fully symmetric (invariant to ordering of the input pair) and thus training
becomes much more efficient. All of the details of this design will be shown in
Subsection \ref{sec:type4impl}. Both of our proposed similarity methods implement
\(G(\cdot)\) as neural networks. The Type 4 measurement design implements
\(C(\hat{\vect{x}},\hat{\vect{y}})\) as a combination of a static binary function
and neural network.

\subsection{Reference similarity measures}
\label{sec:org3a6ab57}
As a reference for our own similarity measure we implemented several reference
similarity measures of Type 1, Type 2 and Type 3. First we
implemented a standard uniformly weighted global similarity (\(t_{1,1}\))
measure which can be defined as:

\begin{equation}
\label{eq:simdef_lin}
t_{1,1}(\vect{x},\vect{y}) = \mathbb{S}(\vect{x},\vect{y}) = C(\vect{x},\vect{y}) = \sum_{i}^{M} \vect{w}_{i} \cdot sim_{i}(\vect{x}_{i},\vect{y}_{i}),
\end{equation}
\sloppy
\noindent where \(sim_{i}(\vect{x}_{i},\vect{y}_{i})\) denotes the local
similarity of the \(i\)-th of \(M\) attributes. In t\textsubscript{1,1} all weights and local
similarity measures are uniformly distributed, and not parametrized by the data.

We extended this with a Type 2 similarity measure \(t_{2,1}\), which is
based on the work from Abdel-Aziz et al. \cite{abdel2014learning}, where the local
similarity measures are parametrized by the data from the corresponding
features.

\begin{figure}[!tbh]
\centering
\includegraphics[width=0.6\linewidth]{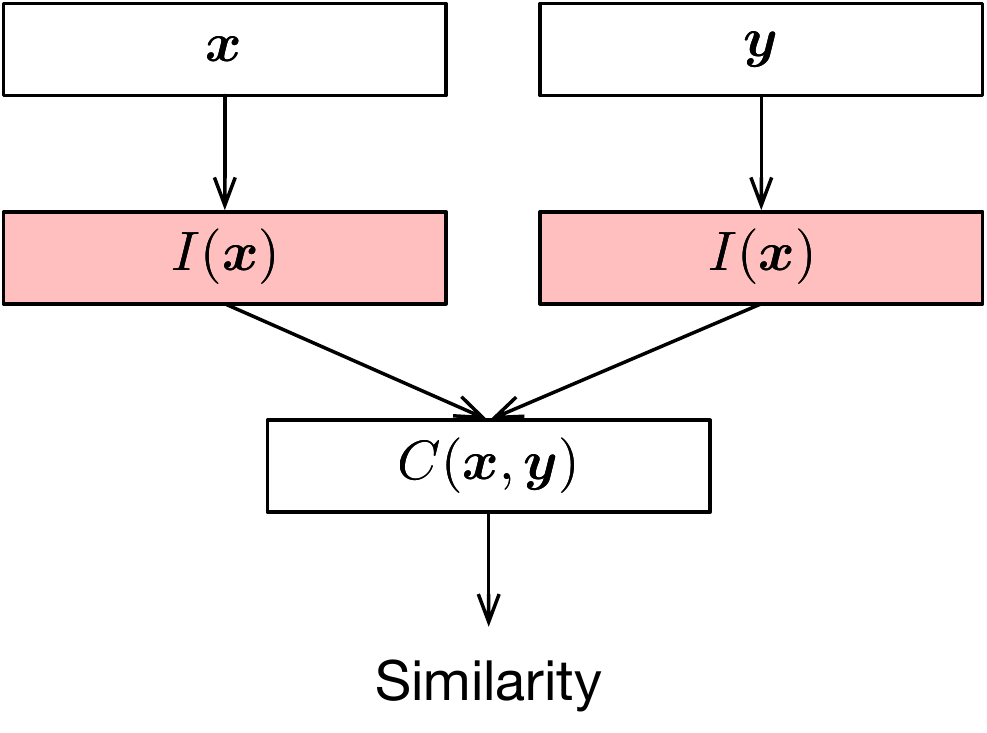}
\caption[gabel2015ann]{\label{fig:ann-gabel}Architecture of a ANN similarity measure as used in Gabel \cite{gabel2015ann} (Type 2), where \(G(\cdot)\) is set to be the identity function \(G(\vect{x}) = I(\vect{x}) = \vect{x}\).}
\end{figure}

Furthermore we implemented a Type 2 similarity measure \(gabel\) as
described by Gabel et al. \cite{gabel2015ann}. The architecture of \(gabel\) can
be seen in Figure \ref{fig:ann-gabel}.

Lastly we implemented the Type 3 similarity measure \(chopra\) described by
Chopra et al. We did not implement the extension done to the contrastive loss
function as seen in \cite{berlemont2015siamese,lefebvre2013learning} as the change
in the training dataset would be too big. This change would make comparisons
between the methods harder to justify. Also none of these works showed any
comparisons with previous SNNs in terms of any increased performance in relation
to regular contrastive loss.

\subsection{Type 3 similarity measure}
\label{sec:orgaf285f4}
\label{sec:type3impl}
In this subsection we will detail how we model the Type 3 similarity
measure \(t_{3,1}\) which uses an embedding function \(G(\cdot)\) trained as a
classifier. This embedding function maps the input point, \(\vect{x}\), to an
embedding space (see Figure \ref{fig:problem-embedding-solution-space}) which
dimensions represents the probabilities of \(\vect{x}\) belonging to a class. We then
model the similarity measure between two points as the a static function
(\(C(\cdot)\) between their two respective embeddings.

\begin{figure}[!tbh]
\centering
\includegraphics[width=0.6\linewidth]{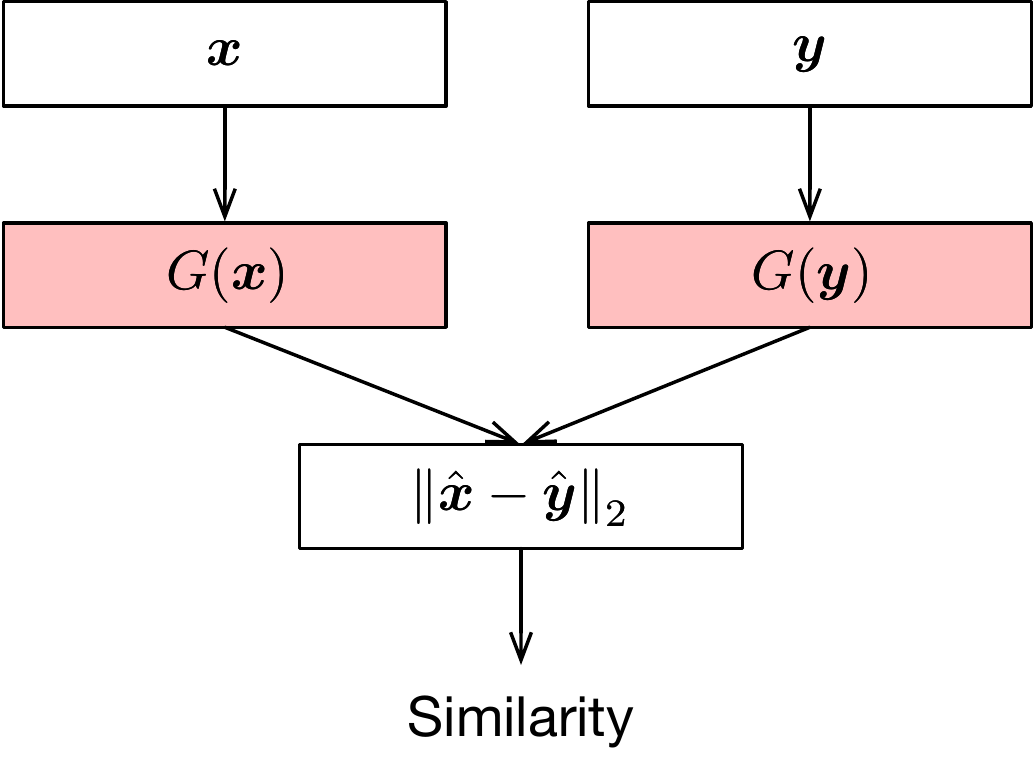}
\caption{\label{fig:t3i1-arch}Architecture of the \(t_{3,1}\) similarity measure where \(G(\cdot)\) is trained to output softmax vectors for classification and the similarity is calculated as a modeled \(L2\) norm between these two vectors (Type 3).}
\end{figure}

For this we choose the \(L2\) norm. So replacing \(C(\cdot)\) for \(L2\) in Equation
\ref{eq:bmmsim}: \(C(\hat{\vect{x}},\hat{\vect{y}}) =
\norm{\hat{\vect{x}} - \hat{\vect{y}}}_{2}\), we can redefine Equation \ref{eq:bmmsim}
to be:

\begin{equation}
\label{eq:t3i1}
 \mathbb{S}(\vect{x},\vect{y}) = t_{3,1}(\vect{x},\vect{y}) = 1.0 - \norm{G(\vect{x}) - G(\vect{y})}_{2}
\end{equation}

\noindent where \(G(\cdot)\) outputs the modeled solution as a \(n\)
dimensional vector (the feature vector output from the network to the softmax
function for \(n\) classes) for the case based on the problem attributes of data
point \(x\). This means that if the \(G(\vect{x})\) evaluates the two cases as very
similar in terms of classification \(G(\vect{x}) \approx G(\vect{y})\) and
\(\norm{G(\vect{x}) - G(\vect{y})} \approx 0\) then \(\mathbb{S}(\vect{x},\vect{y})
\approx 1.0\). This architecture is also illustrated in Figure \ref{fig:t3i1-arch}

Following the model for the \(t_{3,1}\) similarity measure we define the loss
estimate as log-loss between \(G(\vect{x}) = \hat{\vect{x}}\) and \(\vect{t}\),
where \(\vect{t}\) is the is true classification softmax vector, \(\hat{\vect{x}}\)
is the class probability vector output from \(G(\vect{x})\). Notice that the error
estimate of \(t_{3,1}\) does not depend on the output of
\(C(\hat{\vect{x}},\hat{\vect{y}})\).

A data-set of size \(N\) would then be defined as:
\begin{equation}
\label{eq:type3dataset}
\vect{T} = \biggl[(\vect{x}^{1},\vect{t}^{1}) \ldots  (\vect{x}^{N},\vect{t}^{N}) \biggr], 
\end{equation}

\noindent where \(\vect{x}^{N}\) is the problem part of the \(N\)-th data
point and \(\vect{t}^{N}\) is the solution/target part of the \(N\)-th data point.

If the relation between the problem part of the data point (\(\vect{x}\)) and
the solution part of the data point (\(\vect{t}\)) is complex, the network
architecture needs to be able to represent the complexity and any
generalizations of patterns in that complexity. 

\subsection{Type 4 similarity measure}
\label{sec:org68f95a1}
\label{sec:type4impl}
As previously explained, Type 4 similarity measures are currently the most
unexplored type of similarity measure. It is also the type of similarity measure
that requires the least amount of modeling. In principle Type 4 similarity
measures learns two things: \(G(\cdot)\) learns a useful embedding, where the most
useful parts of \(\vect{x}\) and \(\vect{y}\) is encoded into \(\hat{\vect{x}}\) and
\(\hat{\vect{y}}\). \(C(\hat{\vect{x}},\hat{\vect{y}})\) learns how to combine those
embeddings to calculate the similarity of the original \(\vect{x}\) and
\(\vect{y}\).

In Type 4 similarity measures both \(C(\hat{\vect{x}},\hat{\vect{y}})\) and
\(G(\cdot)\) are learned. In our Type 4 similarity method we will use an ANN
to represent both \(G(\cdot)\) and \(C(\hat{\vect{x}},\hat{\vect{y}})\). This has
the advantage that the learning on \(\mathbb{S}(\vect{x},\vect{y})\) is an end to
end process. The loss computed after \(C(\hat{\vect{x}},\hat{\vect{y}})\) can be
used to compute gradients for both \(C(\hat{\vect{x}},\hat{\vect{y}})\) and
\(G(\cdot)\). \(C(\hat{\vect{x}},\hat{\vect{y}})\) will learn the binary combination
best suited to calculate the similarity of the two embeddings, while \(G(\cdot)\)
will learn to embed the two data points optimally for calculating their
similarity in \(C(\hat{\vect{x}},\hat{\vect{y}})\). In principle any ML method
could be used to learn \(G(\cdot)\) and \(C(\hat{\vect{x}},\hat{\vect{y}})\), but
not all ML methods lend themselves naturally to back-propagating the error
signal from \(C(\hat{\vect{x}},\hat{\vect{y}})\) through \(G(\cdot)\) and back to
the input.

\begin{figure}[t]
\centering
\includegraphics[width=0.6\linewidth]{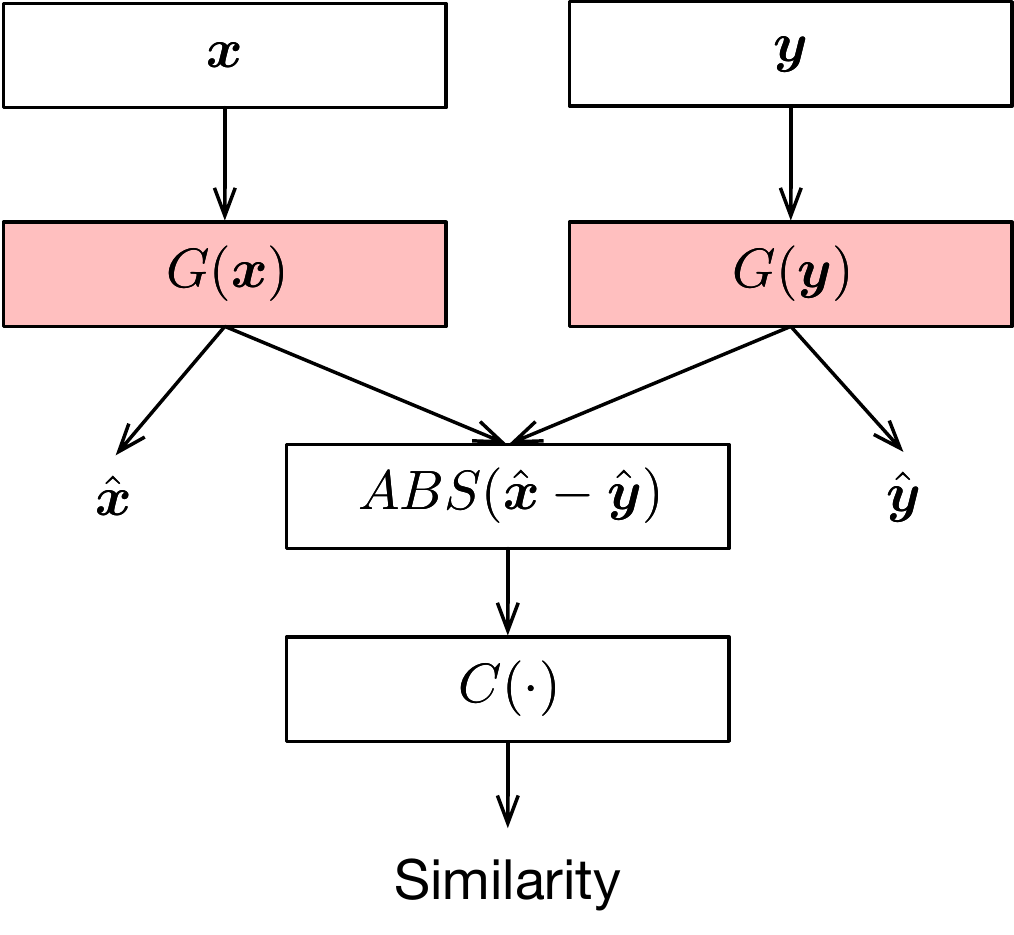}
\caption{\label{fig:eSNN-arch}Architecture of a \(eSNN\) where we combine the symmetry of SNNs with the ability to learn \(C(\hat{\vect{x}},\hat{\vect{y}})\). \(C(\hat{\vect{x}},\hat{\vect{y}})\) is expanded in this picture to highlight the \(ABS(\hat{\vect{x}} - \hat{\vect{y}})\) operation done as the first operation of \(C(\hat{\vect{x}},\hat{\vect{y}})\) to keep \(C\) invariant to the ordering of inputs. It also illustrates the two additional loss signals to \(G(\cdot)\) which helps train the similarity measure.}
\end{figure}

We define our Type 4 similarity method, Extended Siamese Neural Network
(\(eSNN\)) as shown in figure \ref{fig:eSNN-arch}.

Given that this similarity method outputs similarity and the loss function is a
function of the input, we get a new general loss function for similarity,
defined per data-point as follows:

\begin{equation}
\label{eq:t4error}
L_{s}(\vect{x},\vect{y},s) = \lvert s - C(G(\vect{x}),G(\vect{y}))\rvert ,
\end{equation}

\noindent where \(\vect{s}\) is the true similarity of case \(\vect{x}\)
and \(\vect{y}\). Since this loss function is dependent on pairs of data points
and the true similarity between them, we need to create a new dataset based on
the original dataset. This new dataset consists of triplets of two different
data points from the original dataset and the true similarity of these two
data points:

\begin{equation}
\label{eq:type4dataset}
\vect{T} = \biggl[(\vect{x}^{1},\vect{y}^{1},s^{1}) \ldots (\vect{x}^{N},\vect{y}^{N},s^{N}) \biggr], 
\end{equation}

\noindent where \(\vect{s}^{N}\) is \(1\) if \(\vect{x}^{N}\) and
\(\vect{y}^{N}\) belong to the same class and \(0\) otherwise.

It is worth to mention that this dataset is of size \(N(N-1)\) for the similarity
measure to train on all possible combinations of the \(N\) data points. Certain
similarity measure architectures (e.g. \(gabel\) from Gabel et
al.\cite{gabel2015ann} or Zagoruyko et al.'s similarity measures
\cite{zagoruyko2015learning} ) needs to train on a dataset containing all possible
combinations of data points (of size \(N(N-1)\)) as training on the triplet
\((\vect{x},\vect{y},s)\) does not guarantee that the model learns that
\(\mathbb{S}(\vect{y},\vect{x}) = s\). Thus the training dataset must
also include the triplet \((\vect{y},\vect{x},s)\). However this may be largely
avoided by using architectures (such as those seen in SNNs and SNs) that exploit
symmetry and weight sharing. To achieve this we modeled \(C(\vect{x},\vect{y})\)
as a ANN where the first lay.er is an absolute difference operator on two
vectors: \(\vect{z} = ABS(\hat{\vect{x}} - \hat{\vect{y}})\). where \(\vect{z}\) is
the element-wise absolute difference between \(\hat{\vect{x}}\) and
\(\hat{\vect{x}}\). The rest of \(C(\hat{\vect{x}},\hat{\vect{x}})\) is hidden
layers of ANN that operate on \(\vect{z}\). This way
\(C(\hat{\vect{x}},\hat{\vect{x}})\) becomes invariant to the ordering of inputs
to \(S(\vect{x},\vect{y})\). Consequently the model only needs to train on
order-invariant unique pairs of data points, reducing training set size from
\(N(N-1)\) to \(N(\frac{N}{2} - 1)\). The resulting architecture of \(eSNN\) can be seen
in \ref{fig:eSNN-arch}.

In Subsection \ref{sec:type3impl} we argue why \(G(\cdot)\) trained to correctly
classify its input is a good embedding function for calculating similarity.
As a result we added two loss signals to \(G(\cdot)\) during training. These loss
signals are calculated from the difference between the embedding of the data point
produced by \(G(\cdot)\) and the correct soft-max classification vector.

\begin{figure}[t]
\centering
\includegraphics[width=0.9\linewidth]{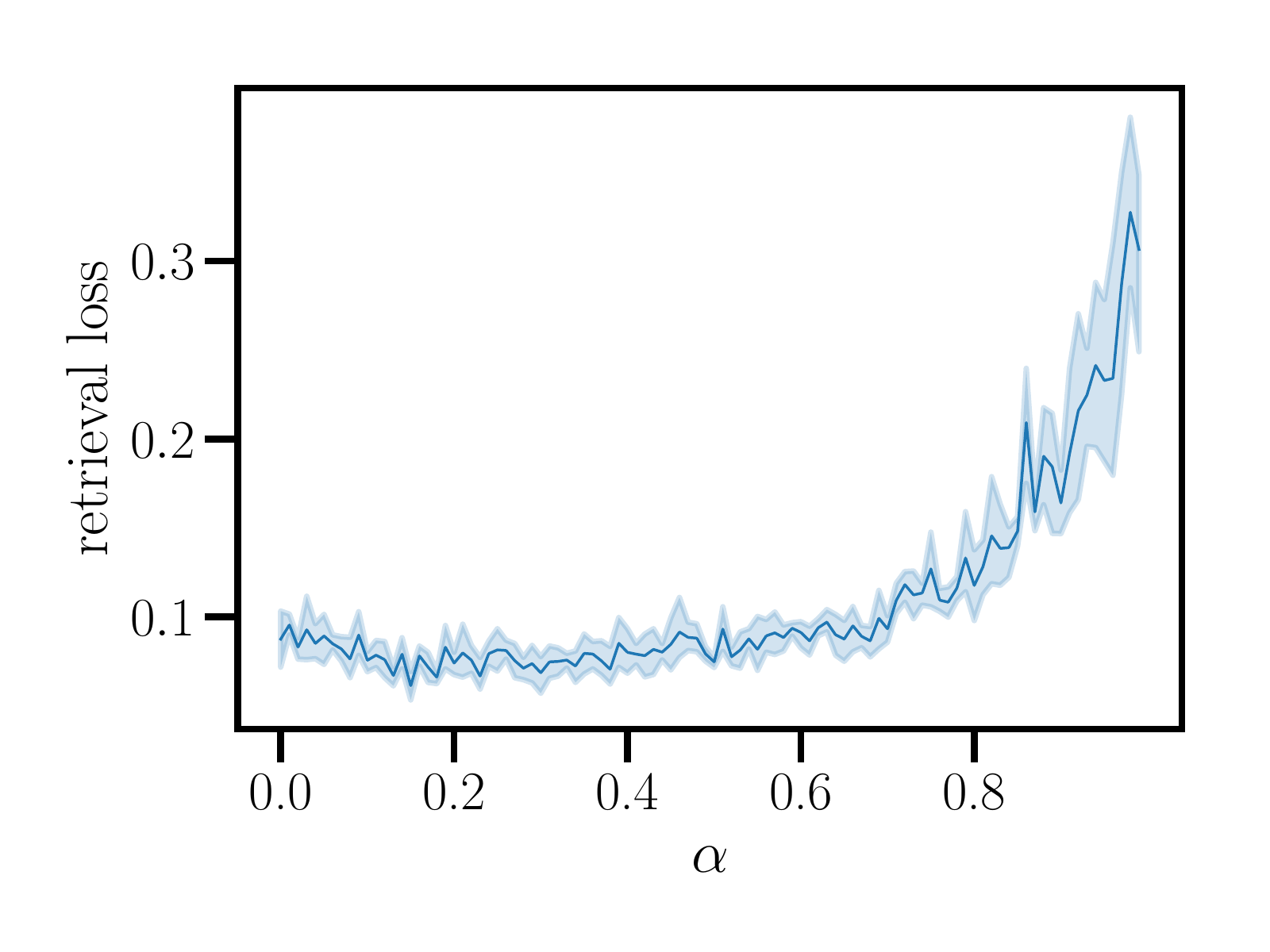}
\caption[dua2017]{\label{fig:alphasearch}Showing results from weighting the three different output in terms of signal strength to loss measured on the UCI dataset balance scale \cite{dua2017} (5-fold cross validation and repeated 5 times). This measurement was done using training data of size \(N(\frac{N}{2} - 1)\). The effect of \(\alpha\) is much less impactful on the validation result after 200 or more epochs of training when training on \(N(N-1)\) datasets. However choosing the correct \(\alpha\) using \(N(\frac{N}{2} - 1)\) datasets does impact the speed of training for \(eSNN\) when training on \(N(N-1)\) datasets.}
\end{figure}

This also introduced an opportunity for exploring the relative importance of the
embedding function \(G(\cdot)\) and the binary similarity function
\(C(\cdot)\) in terms of the performance of the similarity measure.
This could be done by weighting the three different loss signals
(\(\hat{\vect{x}}\), \(\hat{\vect{y}}\) and similarity as shown in Figure
\ref{fig:eSNN-arch}) during training and measuring the effect of that weighting on
the performance. We define our weighted loss function as such:

\begin{align}
\label{eq:alphaeq}
L(\alpha,\vect{x},\vect{y},\vect{s}) &= \frac{(1-\alpha)}{2} \cdot (L_{c}(\vect{x},\vect{t}_{x})+L_{c}(\vect{y},\vect{t}_{y})) \nonumber \\
                            & +\alpha \cdot L_{s}(\vect{x},\vect{y},s),
\end{align}

\noindent where \(L_{s}(\cdot)\) is defined in Equation \ref{eq:t4error},
\(\vect{t}_{x}\) is the true label of data point \(\vect{x}\), \(\vect{t}_{y}\) is the
true label of data point \(\vect{y}\) and \(L_{c}(\vect{v}_{1},\vect{v}_{2})\) is
the categorical cross entropy loss between two softmax vectors. We use this
formula and tested with different 100 different values of \(\alpha\) in the range
\([0,1]\) to find the weighting scheme best for performance. The results can be
seen in Figure \ref{fig:alphasearch}.

Figure \ref{fig:alphasearch} seems to indicate that \(\alpha = 0.15\) is ideal for
this dataset. We have used \(\alpha = 0.15\) throughout the experiments reported
in Section \ref{sec:evaluation}.

\subsection{Network parameters}
\label{sec:org302367c}
For all similarity measures tested using ANN and all datasets except MNIST,
\(G(\cdot)\) and \(C(\cdot)\) where implemented with two hidden layers of \(13\)
nodes. This was done to replicate the network parameters used by Gabel et al. to
ensure we had comparable results. For the MNIST dataset test both \(chopra\)
and \(eSNN\) used three hidden layers of \(128\) nodes for \(G(\cdot)\), and the same
for \(C(\cdot)\)

\begin{figure}[t]
\centering
\includegraphics[width=0.9\linewidth]{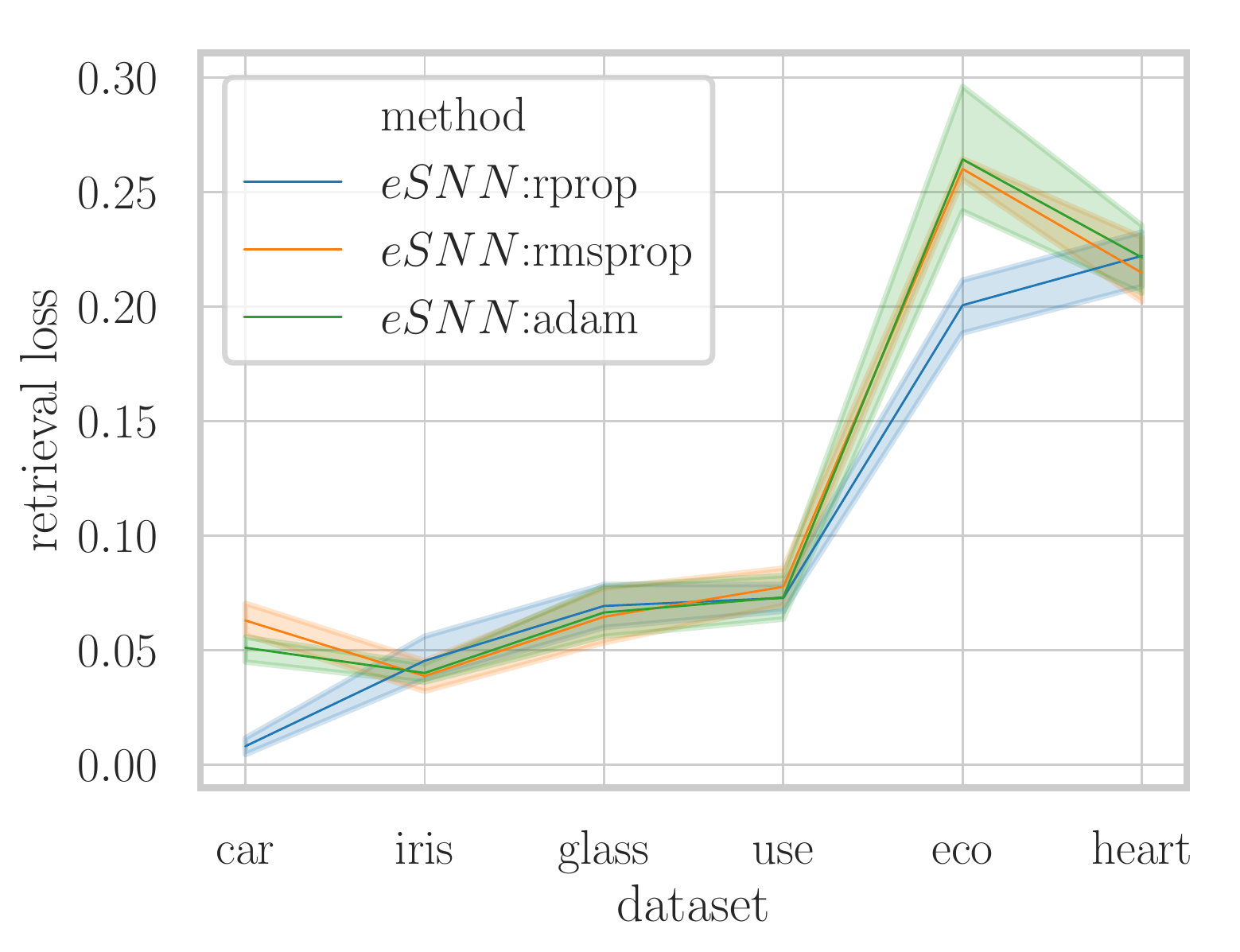}
\caption{\label{fig:rpropeval}Testing how the RProp algorithm performs in comparison with ADAM and RMSProp. Our proposed architecture performs best using the RProp algorithm (5-fold cross validation and repeated 5 times).}
\end{figure}

Other than the network architecture we also wanted to choose which optimizer to
use for learning the ANN model. We wanted to chose the RProp
\cite{riedmiller1993direct} to be more comparable with the results from Gabel et
al. which also used RProp optimizer. Our tests seen in Figure \ref{fig:rpropeval}
shows that RProp outperforms all other optimizer tested (ADAM and RMSProp). This
is consistent with the results reported by Florescu and Igel
\cite{florescu2018resilient}. This should hold true until the the run-time
performance of RProp degrades with dataset size, as RProp uses full batch sizes.

\subsection{Evaluation protocol and implementation}
\label{sec:org2b51817}
\label{sec:protocol}

The different similarity measures presented earlier in this section requires
different training data sets. The reference Type 1 similarity measures
(\(t_{1,1}\)) requires no training. While \(t_{2,1}\) and \(t_{3,1}\) does not
require a similarity training consisting of triplets as described in Equations
\ref{eq:type4dataset}. All other similarity measures evaluated was trained using
identical training datasets. As a result, all similarity measures were trained
on a dataset consisting of all possible combinations of data points (as explained
in \ref{sec:type4impl}) as this is required by the \(gabel\) similarity measure.
However, results highlighting the differences in training performance when using
the different training datasets can be seen in Figure \ref{fig:trainingloss2}.

The results reported in the next section are all 5-fold stratified cross
validation repeated 5 times for robustness. The performance reported is an
evaluation of each similarity measurement using the part of the dataset
(validation partition) that was not used for training. Using the similarity
measure being evaluated, we computed the similarity between every data point in
the validation partition (\(V\)) and every data point in the training partition
(\(T\)). For each validation data point (\(x_{v} \in V\)) we find the data
point in the training set \(T\) with the highest similarity (\(x_{t} =
\argmax\nolimits_{x_{i} \in T}(\mathbb{S}(x_{v},x_{i}))\)). If \(x_{t}\) has the
same class as \(x_{v}\) from the validation partition, we scored it as \(1.0\), if
not, we scored it as \(0.0\).

The implementation was done in Keras \footnote{Code available at NTNU OpenAI lab
github page: \url{https://github.com/ntnu-ai-lab}} with Tensorflow as backend. The
methods was measured on 14 different datasets available from the UCI machine
learning repository \cite{dua2017}. Results was recorded after 200 epochs and 2000
epochs (the latter number to be consistent with Gabel et al. \cite{gabel2015ann})
to reveal how fast the different methods were achieving their performance. 

\section{Experimental evaluation}
\label{sec:orgd05f60c}
\label{sec:evaluation}
To calculate the performance of our similarity measure we chose to use the same
method of evaluation as Gabel et al. \cite{gabel2015ann} to make the similarity
metrics more easily comparable. In addition this evaluation method of using
publicly available datasets from the UCI machine learning repository
\cite{dua2017} make the results easy to reproduce. We selected a subset of the
original 19 datasets, choosing not to use regression datasets, resulting in a
set of 14 classification datasets. The datasets' numerical features were all
normalized, categorical features were replaced by a one-hot vector.

\begin{figure*}[tp]
\centering
\includegraphics[width=0.8\linewidth]{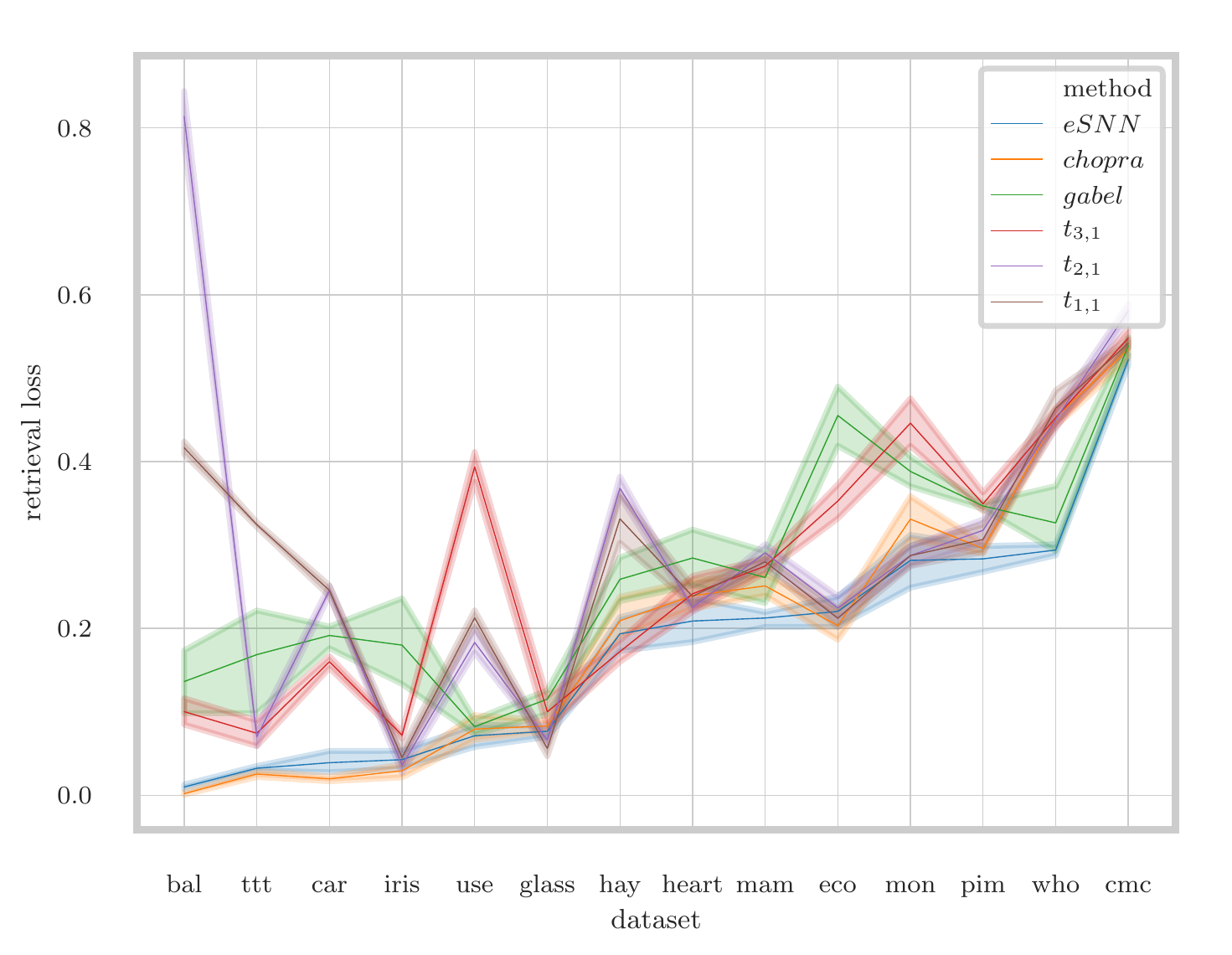}
\caption{\label{fig:200e}Performance of \(eSNN\) in comparison to reference similarity measures and state of the art similarity methods over all test datasets trained over 200 epochs.}
\end{figure*}

\begin{figure*}[tp]
\centering
\includegraphics[width=0.8\linewidth]{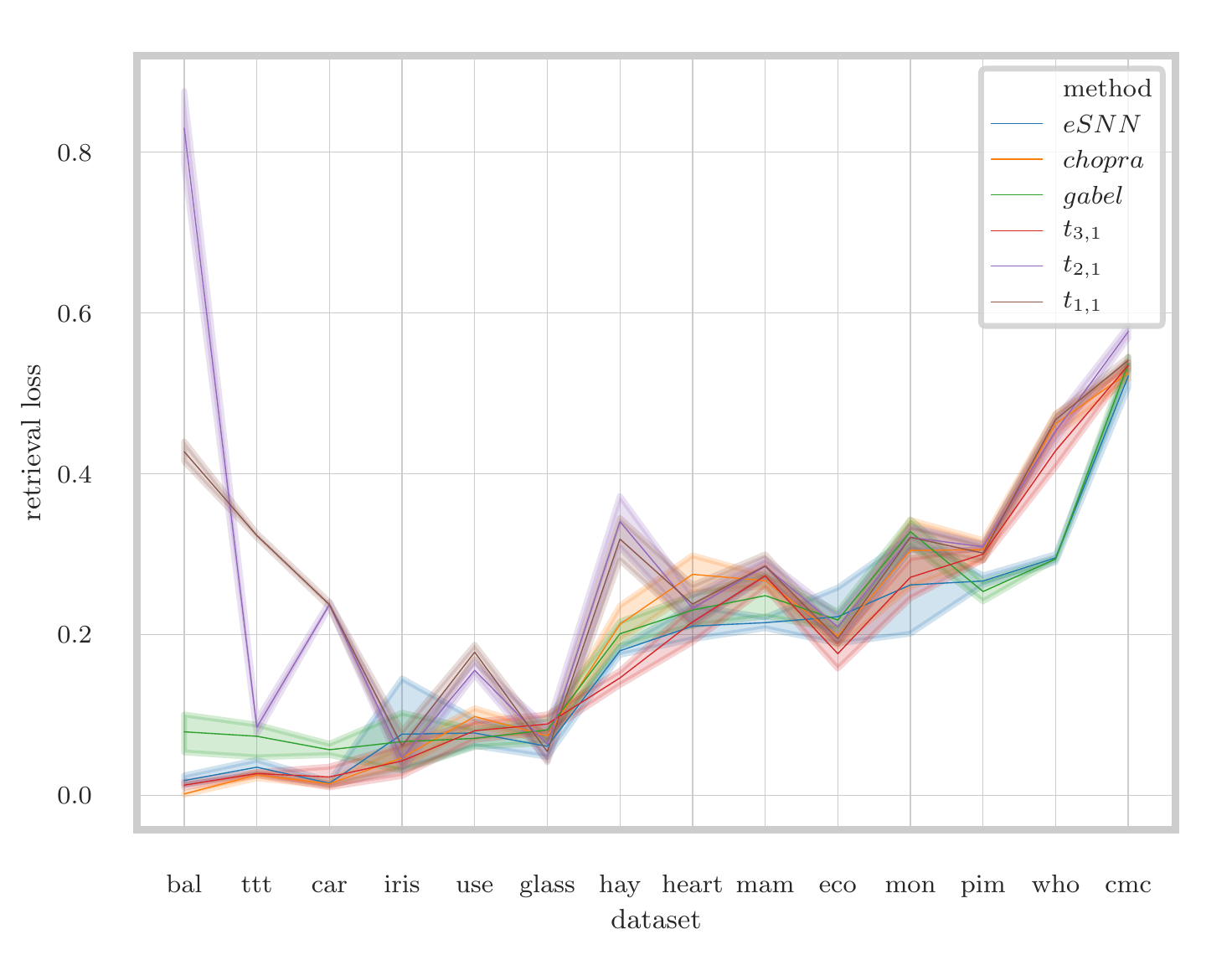}
\caption{\label{fig:2000e}Performance of \(eSNN\) in comparison to reference similarity measures and state of the art similarity methods over all test datasets trained over 2000 epochs.}
\end{figure*}

The validation losses from evaluating the similarity measures on the 14 datasets
are shown in Figures \ref{fig:200e} and \ref{fig:2000e}. Figure \ref{fig:200e} shows
the results after training for 200 epochs, while Figure \ref{fig:2000e} shows the
results after 2000 epochs. This has been done to illustrate how the differences
between the similarity measures develop during training. In addition the \(200\)
and \(2000\) epoch runs are independent runs (i.e. Figure \ref{fig:2000e} is not the
same models as seen in Figure \ref{fig:200e} \(1800\) epochs later)

The numbers that are the basis of these figures are also reported in Table
\ref{tab:tabresults200e} for 200 epochs and Table \ref{tab:tabresults2000e} for 2000
epochs. The tables are highlighted to show the best result per dataset. In some
cases the differences between two methods for one dataset was smaller than the
standard deviation thus highlighting more than one result.

\begin{figure}[tp]
\centering
\includegraphics[width=1\linewidth]{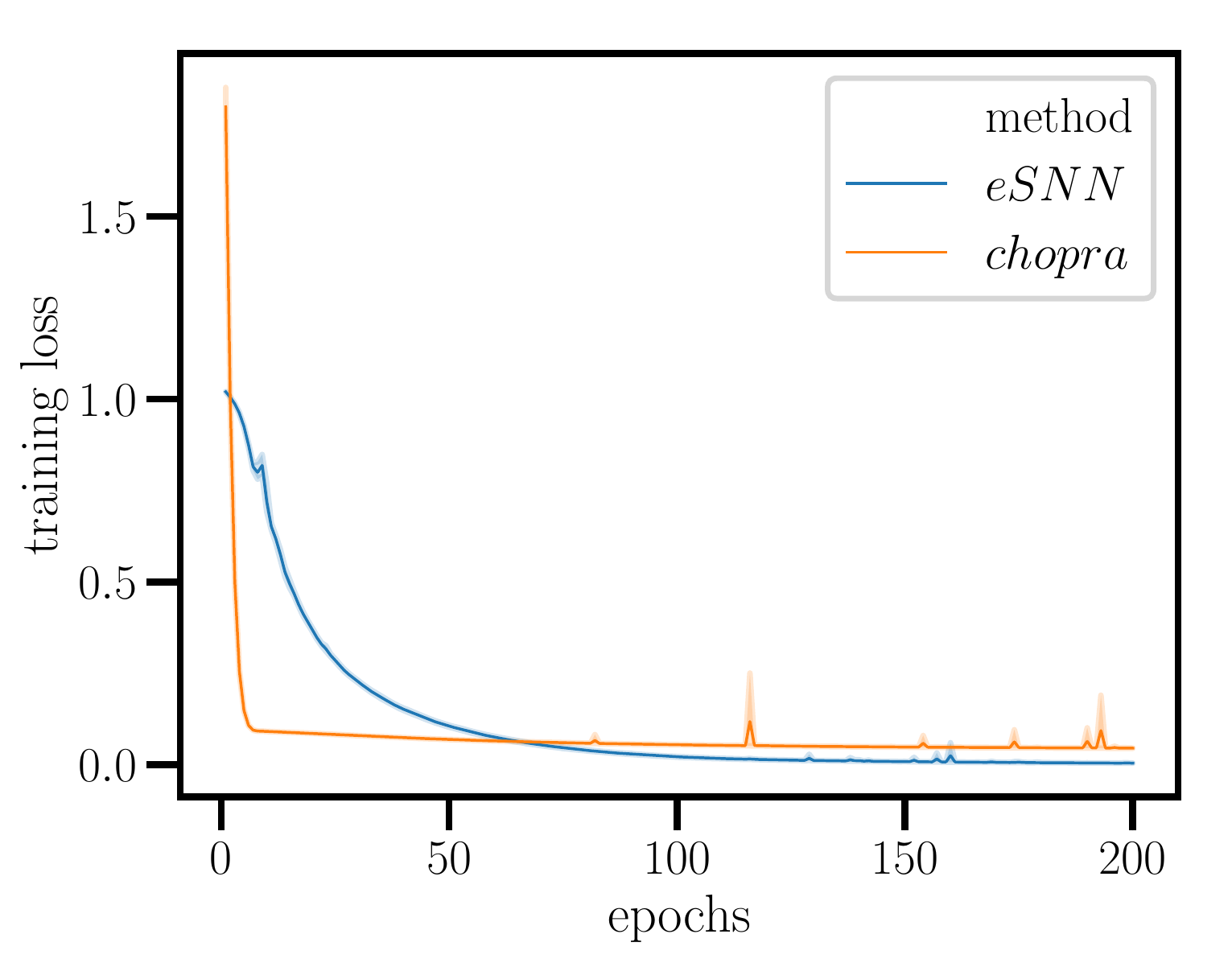}
\caption{\label{fig:mnisttraining}Training loss (not validation retrieval loss) during training on the MNIST dataset for \(chopra\) and \(eSNN\). \(gabel\) could not be evaluated as training on a \(N(N-1)\) sized dataset for MNIST is too resource demanding.}
\end{figure}

Finally, to illustrate that \(eSNN\) scales to larger datasets we report
results from the MNIST dataset in Figure \ref{fig:mnisttraining}. The MNIST
results are not validation results, as calculating the similarity between all
the data points in the test set and the training set (as per the evaluation
protocol described in Section \ref{sec:protocol}) was too resource demanding.

\begin{table}[htbp]
\centering
\begin{tabular}{|l|l|l|l|l|l|l|}
\hline
 & \(eSNN\) & \(chopra\) & \(gabel\) & \(t_{3,1}\) & \(t_{1,1}\) & \(t_{2,1}\)\\
\hline
\hline
bal & 0.01 & \textbf{0.00} & 0.14 & 0.10 & 0.42 & 0.81\\
car & 0.04 & \textbf{0.02} & 0.19 & 0.16 & 0.25 & 0.25\\
cmc & \textbf{0.52} & 0.53 & 0.54 & 0.55 & 0.54 & 0.58\\
eco & 0.22 & \textbf{0.20} & 0.46 & 0.35 & 0.21 & 0.22\\
glass & 0.08 & 0.08 & 0.12 & 0.10 & \textbf{0.06} & 0.07\\
hay & 0.19 & 0.21 & 0.26 & \textbf{0.17} & 0.33 & 0.37\\
heart & \textbf{0.21} & 0.24 & 0.28 & 0.24 & 0.24 & 0.23\\
iris & 0.04 & \textbf{0.03} & 0.18 & 0.07 & 0.05 & 0.04\\
mam & \textbf{0.21} & 0.25 & 0.26 & 0.27 & 0.28 & 0.29\\
mon & \textbf{0.28} & 0.33 & 0.39 & 0.45 & 0.29 & 0.29\\
pim & \textbf{0.28} & 0.30 & 0.35 & 0.35 & 0.31 & 0.32\\
ttt & \textbf{0.03} & 0.03 & 0.17 & 0.07 & 0.32 & 0.07\\
use & \textbf{0.07} & 0.08 & 0.08 & 0.39 & 0.21 & 0.18\\
who & \textbf{0.29} & 0.45 & 0.33 & 0.45 & 0.46 & 0.45\\
\hline
Sum & \textbf{2.47} & 2.75 & 3.75 & 3.72 & 3.97 & 4.17\\
\hline
Average & \textbf{0.18} & 0.20 & 0.27 & 0.27 & 0.28 & 0.30\\
\hline
\end{tabular}
\caption{\label{tab:tabresults200e}Validation retrieval loss after 200 epochs of training, in comparison to state of the art methods. \(eSNN\) has the smallest loss in \(8\) of \(14\) datasets. The best result for each dataset is highlighted in bold.}

\end{table}

Table \ref{tab:tabresults200e} shows the validation losses of the different
similarity measures on the different datasets. Our proposed Type 4
similarity measure \(eSNN\) has \(11\%\) less validation loss than the second
best (Type 3) similarity measure \(chopra\) (Chopra et al.
\cite{chopra2005learning}). The other Type 3 similarity measures follow with
\(t_{3,1}\) having \(51\%\) higher loss and \(gabel\) (Gabel et al.
\cite{gabel2015ann}) with \(52\%\) more loss. The Type 1 similarity measure had
\(61\%\) more loss but managed to be the best similarity measure for the glass
dataset. At last Type 2 similarity measure had \(69\%\) higher loss than
\(eSNN\) on average.

\begin{table}[htbp]
\centering
\begin{tabular}{|l|l|l|l|l|l|l|}
\hline
 & \(eSNN\) & \(chopra\) & \(gabel\) & \(t_{3,1}\) & t\textsubscript{1,1} & \(t_{2,1}\)\\
\hline
\hline
bal & 0.02 & \textbf{0.00} & 0.08 & 0.01 & 0.43 & 0.83\\
car & \textbf{0.01} & \textbf{0.01} & 0.06 & 0.02 & 0.24 & 0.24\\
cmc & \textbf{0.52} & 0.53 & 0.54 & 0.53 & 0.54 & 0.58\\
eco & 0.22 & 0.20 & 0.22 & \textbf{0.18} & 0.19 & 0.21\\
glass & 0.06 & 0.07 & 0.08 & 0.09 & \textbf{0.05} & 0.06\\
hay & 0.18 & 0.21 & 0.20 & \textbf{0.15} & 0.32 & 0.34\\
heart & \textbf{0.21} & 0.27 & 0.23 & 0.22 & 0.24 & 0.23\\
iris & 0.08 & 0.05 & 0.07 & \textbf{0.04} & 0.06 & 0.05\\
mam & \textbf{0.21} & 0.27 & 0.25 & 0.27 & 0.29 & 0.28\\
mon & \textbf{0.26} & 0.30 & 0.33 & 0.27 & 0.32 & 0.32\\
pim & 0.27 & 0.31 & \textbf{0.25} & 0.30 & 0.30 & 0.31\\
ttt & \textbf{0.03} & \textbf{0.03} & 0.07 & \textbf{0.03} & 0.32 & 0.08\\
use & 0.08 & 0.10 & \textbf{0.07} & 0.08 & 0.18 & 0.16\\
who & 0.30 & 0.46 & \textbf{0.29} & 0.43 & 0.47 & 0.45\\
\hline
Sum & \textbf{2.45} & 2.81 & 2.74 & 2.62 & 3.95 & 4.14\\
\hline
Average & \textbf{0.18} & 0.20 & 0.20 & 0.19 & 0.28 & 0.30\\
\hline
\end{tabular}
\caption{\label{tab:tabresults2000e}Validation retrieval loss after 2000 epochs of training, in comparison to state of the art methods. \(eSNN\) has the smallest validation retrieval loss in \(6\) of \(14\) datasets in addition to the lowest average loss. The best result for each dataset is highlighted in bold.}

\end{table}

The results when training for \(2000\) epochs are quite different from those at
\(200\) epochs, as seen by how much closer the other similarity measures are in
Figure \ref{fig:2000e} than in Figure \ref{fig:200e}. \(eSNN\) still outperforms
all other similarity measures on average, but the second best similarity measure
\(t_{3,1}\) is much closer with just \(6.9\%\) higher loss. \(gabel\) is \(11.8\%\)
worse, \(chopra\) is \(14.7\%\) worse, t\textsubscript{1,1} is \(61.2\%\) worse and finally
\(t_{2,1}\) is \(69\%\) worse than \(eSNN\).

The gap between \(eSNN\) and the state of the art is considerable at \(200\) epochs.
This gap shrinks from \(11\%\) at \(200\) epochs to \(6.9\%\) at \(2000\) epochs, which is
still a considerable difference.

\begin{figure}[tp]
\centering
\includegraphics[width=1\linewidth]{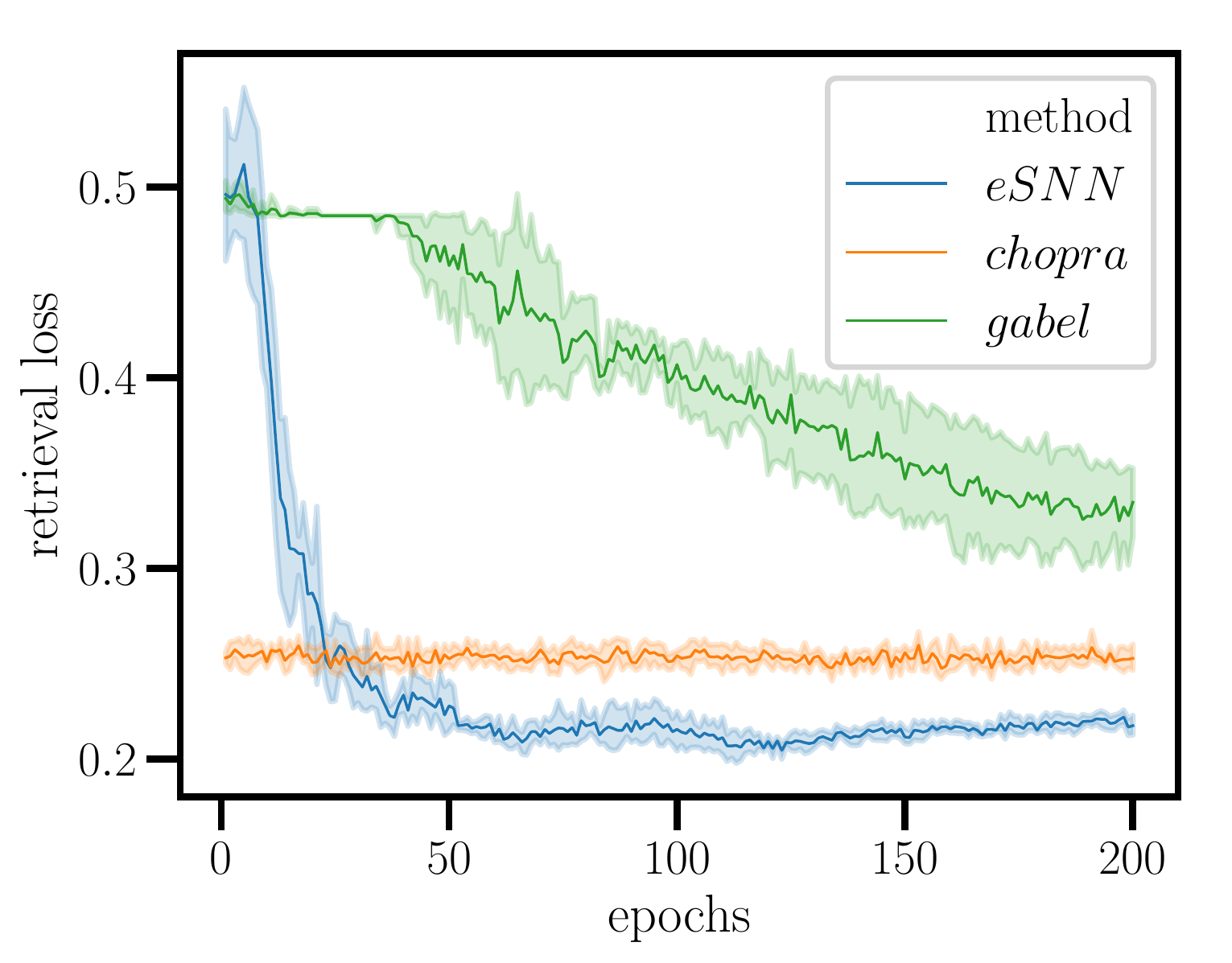}
\caption[dua2017]{\label{fig:trainingloss-mam}Validation retrieval loss during training on the Mammographic mass UCI ML dataset \cite{dua2017}. The Figure shows that the mammograph dataset is a dataset that needs learning outside of embedding via \(G(\cdot)\). \(chopra\) starts out good as \(C(\hat{\vect{x}},\hat{\vect{x}})\) is already designed as the \(L1\) norm. However \(eSNN\) and \(gabel\) catches up when it learns an equivalent and better \(C(\hat{\vect{x}},\hat{\vect{x}})\) function.}
\end{figure}

\begin{figure}[tp]
\centering
\includegraphics[width=1\linewidth]{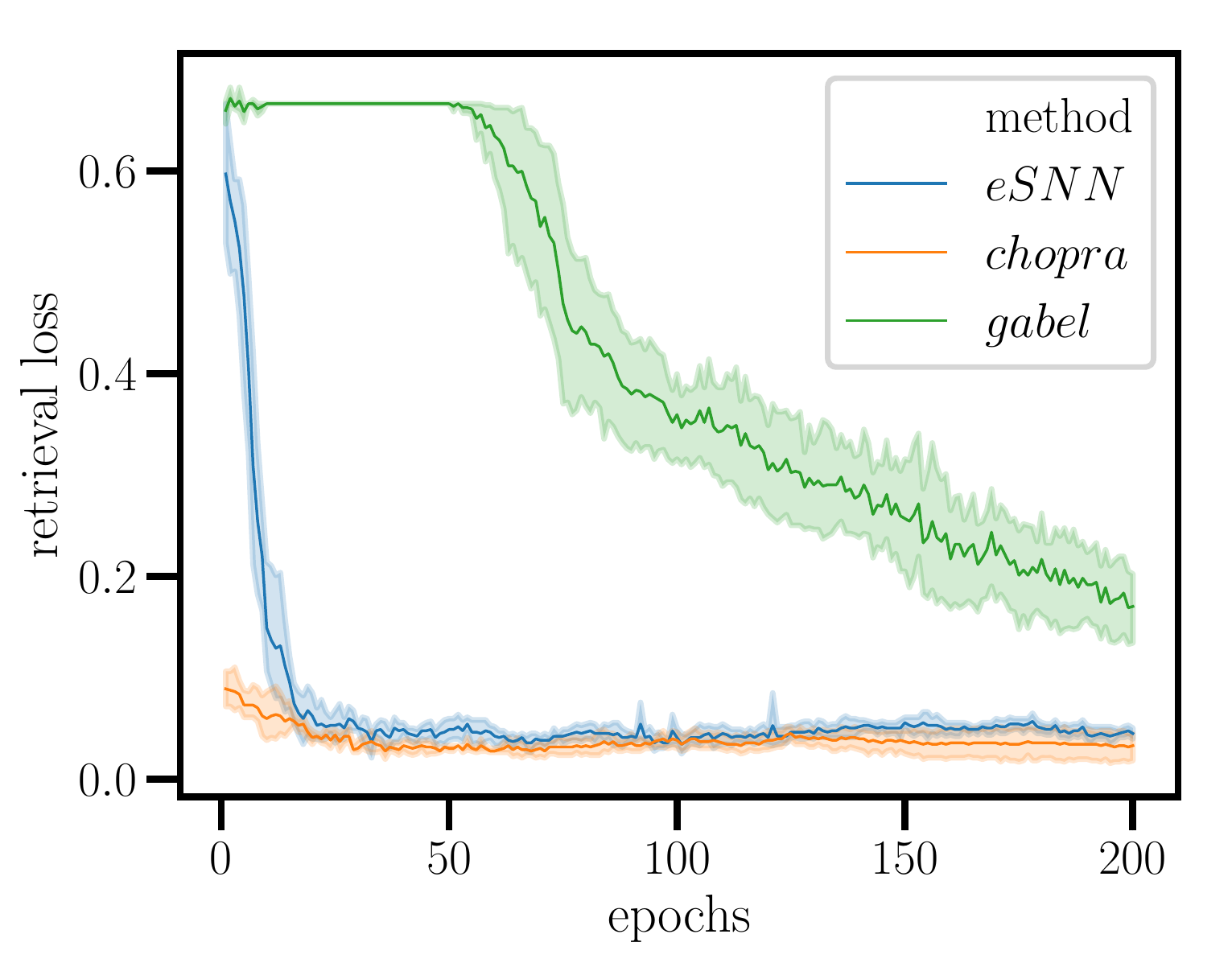}
\caption[dua2017 tab:tabresults2000e]{\label{fig:trainingloss-iris}Validation retrieval loss during training on the Iris UCI ML dataset \cite{dua2017} . Since \(chopra\) starts out with very low validation loss. It seems probable that the static \(L1\) norm \(C(\hat{\vect{x}},\hat{\vect{x}})\) used by \(chopra\) is close to optimal for correctly identifying if the two data points belong to the same class or not. The performance increase done by \(chopra\) is a slight optimization of \(G(\cdot)\). The performance increase done during training by \(gabel\) and \(eSNN\) is mainly by learning a \(C(\hat{\vect{x}},\hat{\vect{x}})\) equivalent in function to that used by \(chopra\), and secondary a slight optimization of \(G(\cdot)\). \(eSNN\) catches up to \(chopra\) in performance after around 20 epochs, however gabel takes longer (5\% validation loss at 2000 epochs) as shown in Table \ref{tab:tabresults2000e}}
\end{figure}

To illustrate the difference in terms of training efficiency between different
types similarity measure, we show the validation loss for \(gabel\),
\(chopra\) and \(eSNN\) during training. Specifically, for each epoch we
test the loss of each similarity measure by the same method as described in
subsection \ref{sec:protocol}. Figure \ref{fig:trainingloss-mam} and Figure
\ref{fig:trainingloss-iris} shows validation loss during training of \(eSNN\),
\(chopra\) and \(gabel\) on the UCI Iris and Mammographic mass datasets
\cite{dua2017} respectively. This exemplifies the training performance of these
methods in relation to the Iris and Mammographic mass dataset results reported
in the tables above. One can also note that in training for the Mammographics
dataset as seen in Fig. \ref{fig:trainingloss-mam} \(chopra\) never achieves the
same performance as \(eSNN\). In contrast, while training on the Iris dataset
(as seen in Fig. \ref{fig:trainingloss-iris}), which is a less complex dataset
than the Mammographic dataset, \(chopra\) achieves the same performance as
\(eSNN\).

\begin{figure}[tp]
\centering
\includegraphics[width=1\linewidth]{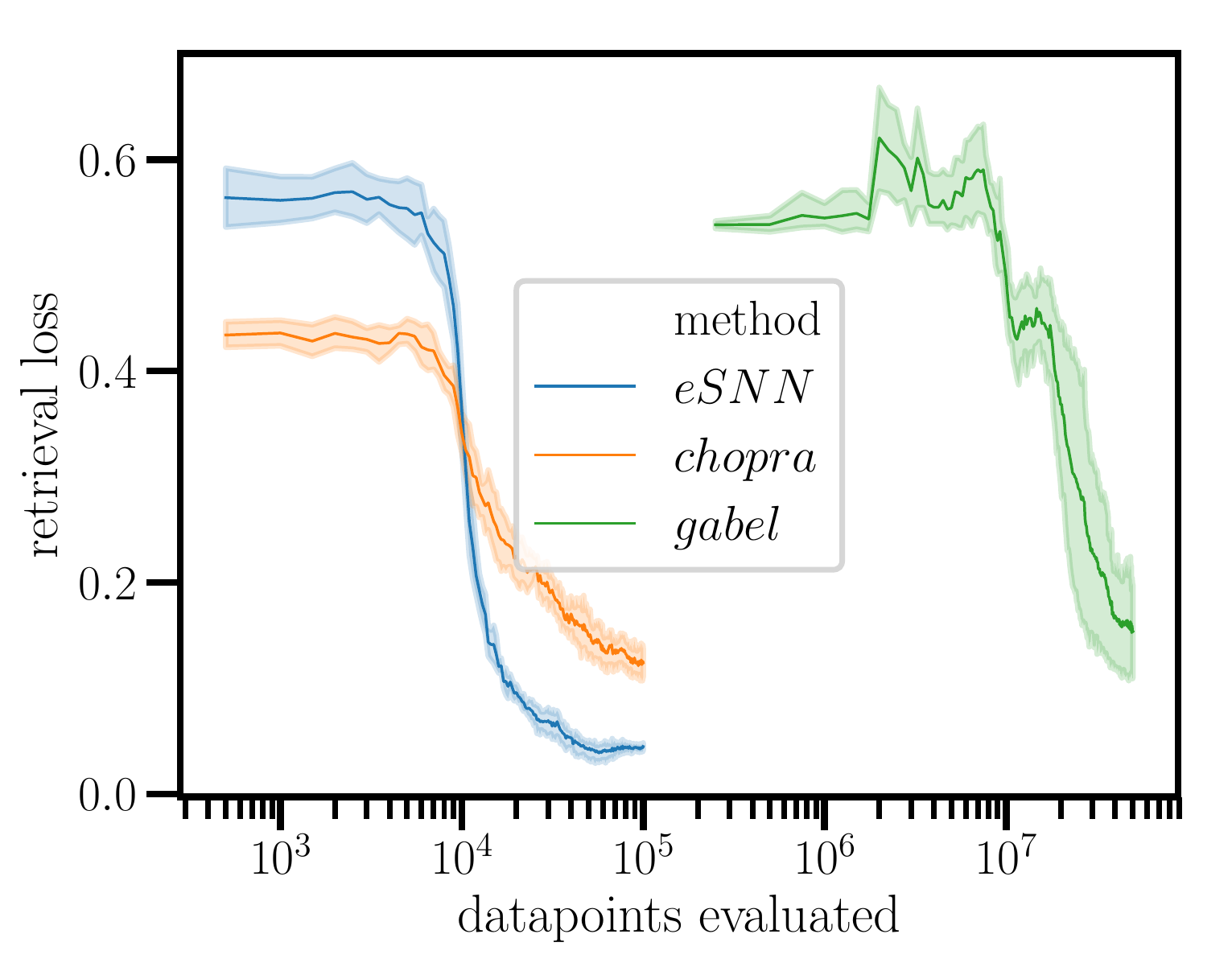}
\caption{\label{fig:trainingloss2}Validation retrieval loss during training on the balance dataset, which illustrates the difference in amount of evaluations needed to achieve acceptable performance. Chopra achieves good performance very quickly, but is outperformed by \(eSNN\) soon. Both have very good performance before having evaluated less (\(N\)) data points than used by one epoch needed by gabel (\(N(N-1)\))}
\end{figure}

Figure \ref{fig:trainingloss2} shows the validation loss during training when
\(chopra\) and \(eSNN\) are using a training dataset of size \(N\) and \(gabel\) is using
a training dataset of size \(N(N-1)\). This figure illustrates how much fewer
evaluations a SNN similarity measure like \(chopra\) or symmetric Type 4
similarity measure such as \(eSNN\) needs than a similarity measurement that is not
invariant to input ordering, while still having excellent relative performance.

\begin{figure}[htbp]
\centering
\begin{subfigure}[c]{0.4\textwidth}
\includegraphics[width=.9\linewidth]{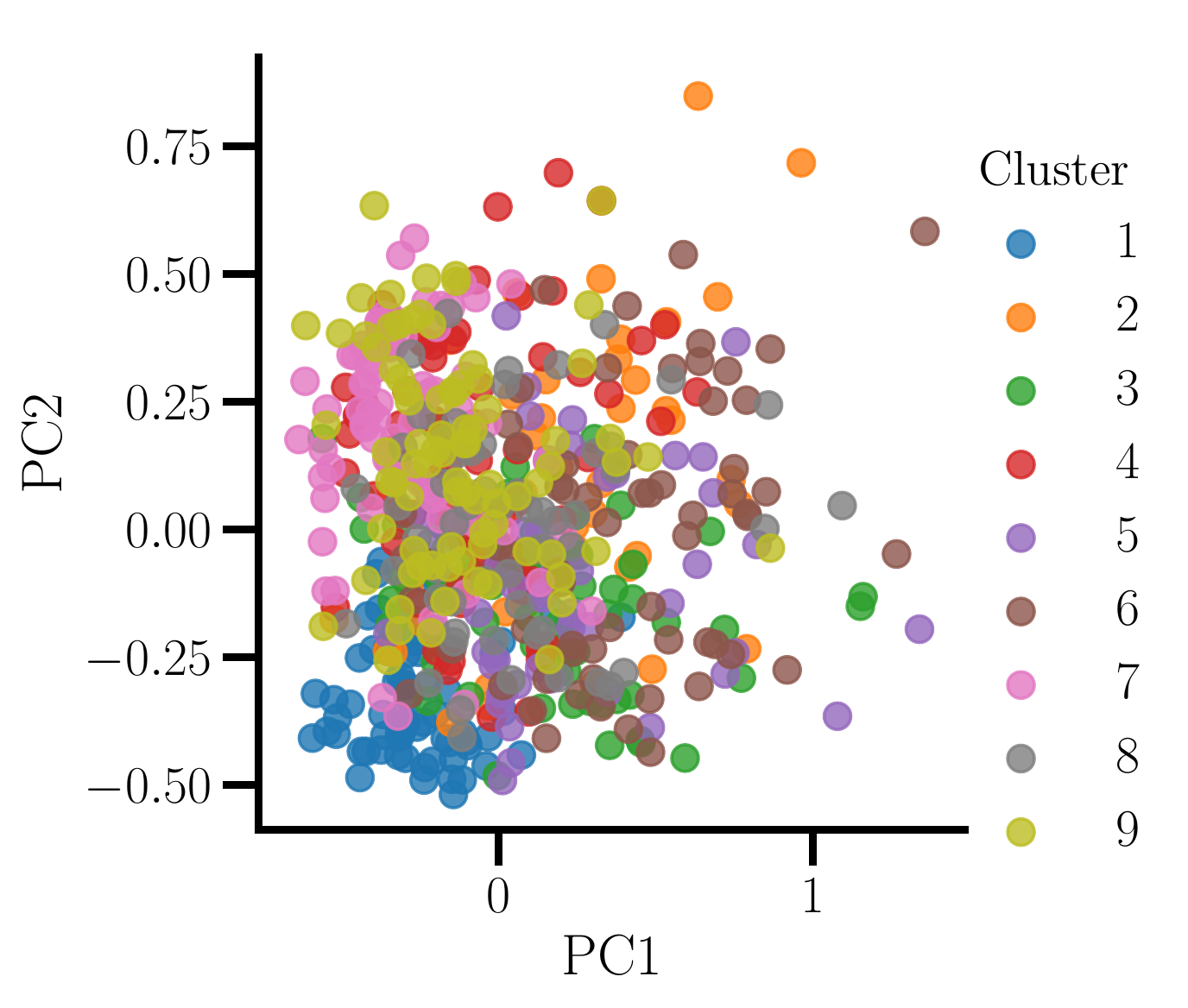}
\caption{\label{fig:pcabeforetraining} PCA clustering on the MNIST dataset before training}
\end{subfigure}
\begin{subfigure}[c]{0.4\textwidth}
\includegraphics[width=.9\linewidth]{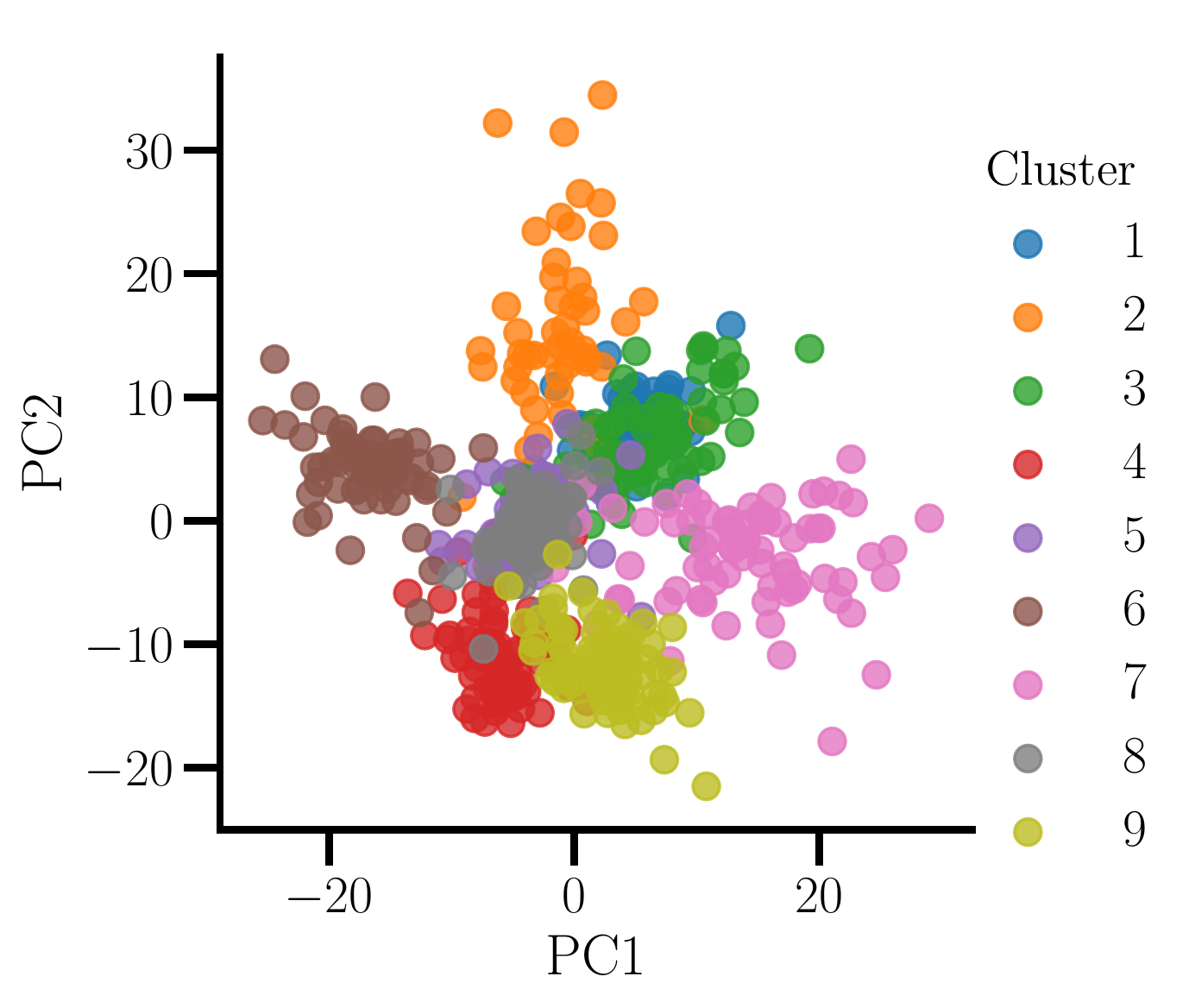}
\caption{\label{fig:pcaaftertraining} PCA clustering on the MNIST dataset after training}
\end{subfigure}
\caption[fig:pcaaftertraining]{\label{fig:mnistembeddings1}PCA clustering showing the two first principal components (\(PCA1\) and \(PCA2\)) of the embeddings produced by \(eSNN\) from MNIST input before (\ref{fig:pcabeforetraining}) and after (\ref{fig:pcaaftertraining}) training.}

\end{figure}

\begin{figure}[htbp]
\centering
\begin{subfigure}[c]{0.4\textwidth}
\includegraphics[width=.9\linewidth]{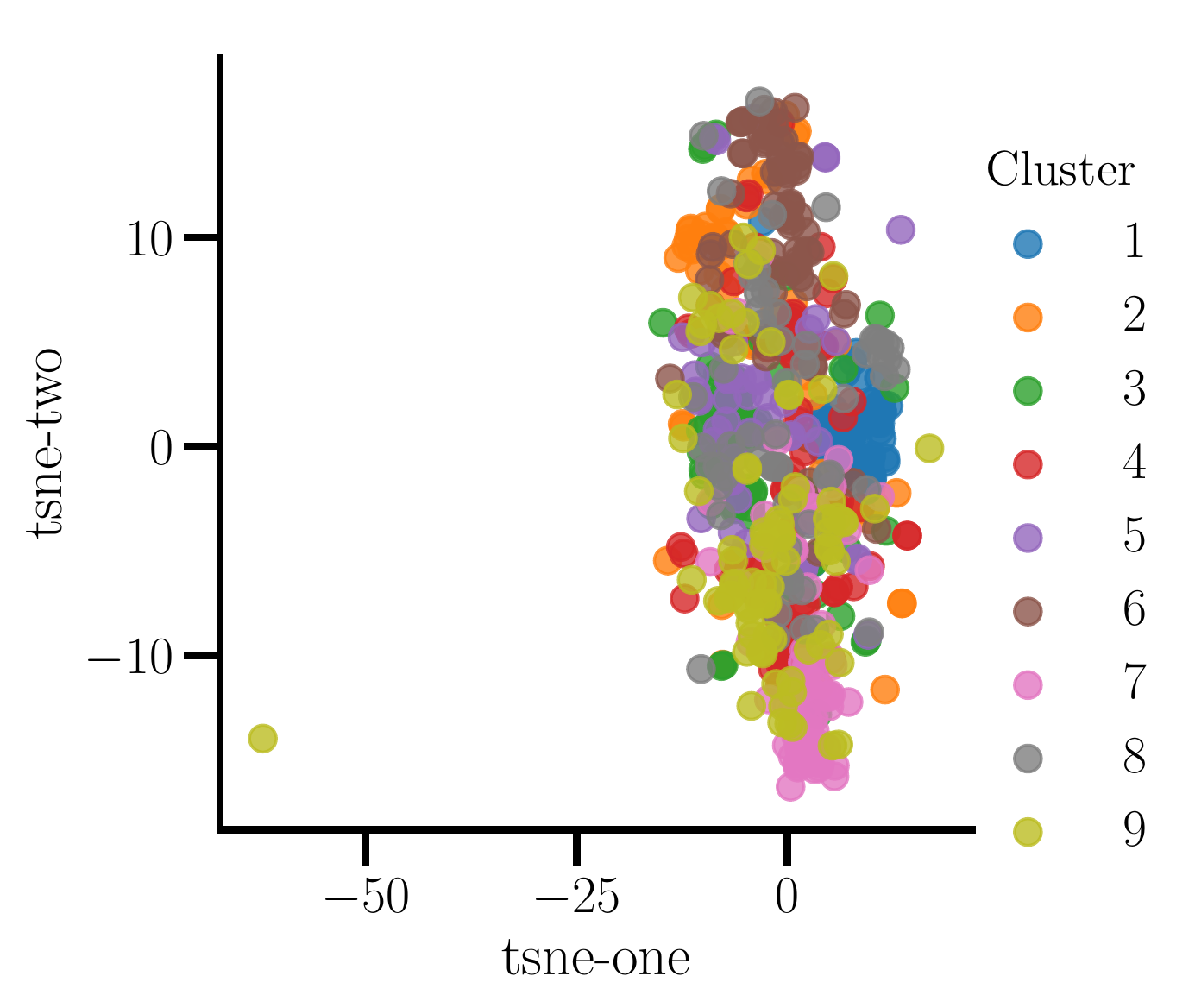}
\caption{\label{fig:t-snebeforetraining} T-SNE clustering of the MNIST dataset before training}
\end{subfigure}
\begin{subfigure}[c]{0.4\textwidth}
\includegraphics[width=.9\linewidth]{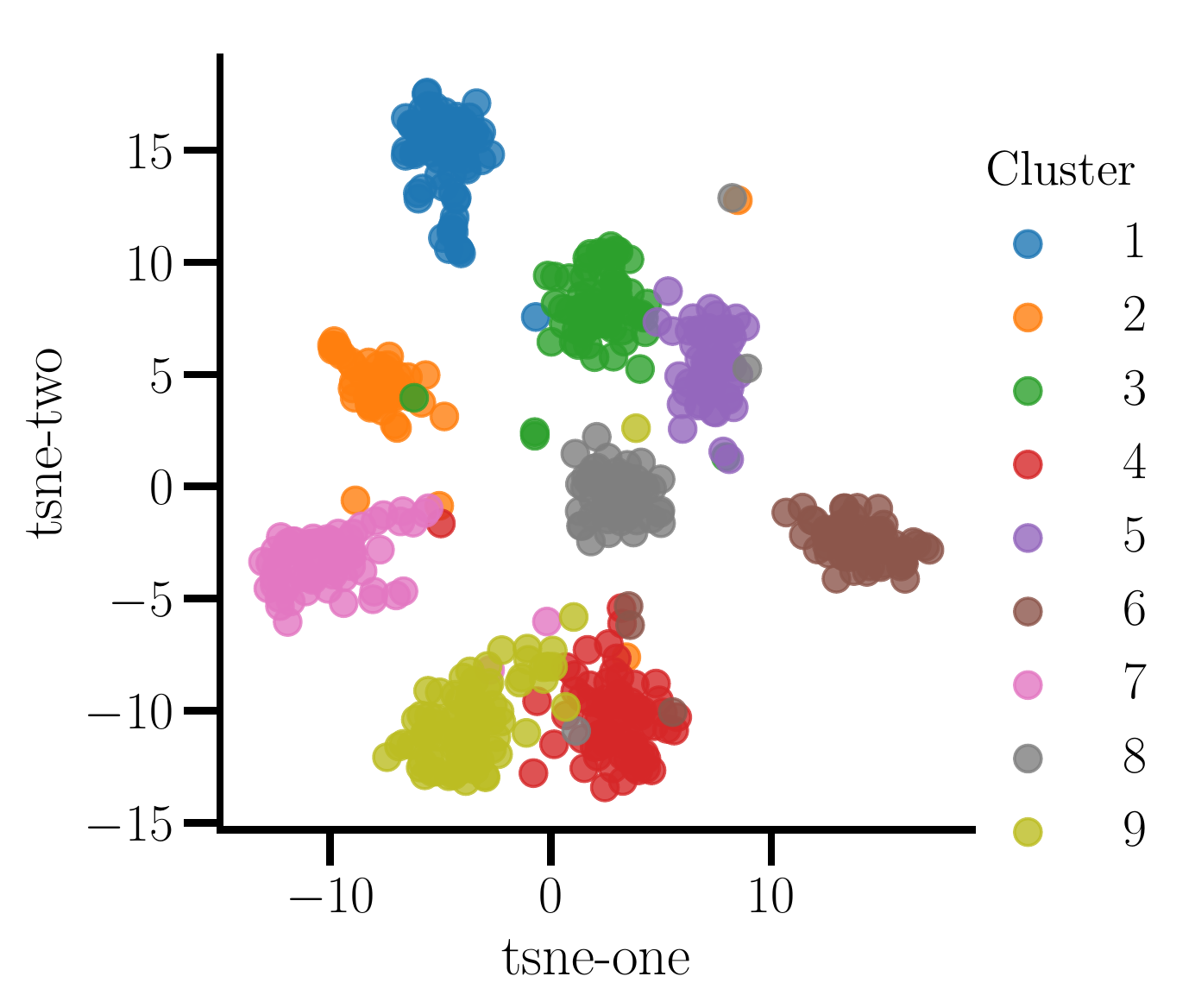}
\caption{\label{fig:t-sneaftertraining} T-SNE clustering of the MNIST dataset after training}
\end{subfigure}
\caption[fig:t-sneaftertraining]{\label{fig:mnistembeddings2}T-SNE clustering of embeddings produced by \(eSNN\) from MNIST input before (\ref{fig:t-snebeforetraining}) and after (\ref{fig:t-sneaftertraining}) training.}

\end{figure}

Finally in Figure \ref{fig:mnistembeddings1} and \ref{fig:mnistembeddings2} we show
how \(eSNN\) can be used for semi-supervised clustering. The figures show PCA and
T-SNE clustering of embeddings produced untrained and trained \(eSNN\) networks
respectively from the MNIST dataset. The embeddings are the vector output of
\(G(\cdot)\) for each of the data points in the test set. The embeddings shown are
computed from a test set that is not used for training. The figures show that
\(eSNN\) learns a way to correctly cluster data points that it has not used for
training.

\section{Conclusions and future work}
\label{sec:org7905d8f}
\label{sec:conclusion}
Section \ref{sec:evaluation} shows that all of the learned similarity measures
outperformed the classical similarity measure \(t_{1,1}\) and also \(t_{2,1}\)
where the local (per feature) similarity measures were adapted to the
statistical properties of the features \cite{abdel2014learning}. In practice one
should weight the importance of each feature according to how important it is in
terms of similarity measurement. In many situations the number of possible
attributes to include in such a function can be overwhelming, and modeling them
in the way we did in \(t_{1,1}\) and \(t_{3,1}\) also overlooks possible
co-variations between the attributes. Both of these problems can be addressed
using the proposed method to model the similarity using machine learning on a
dataset that maps from case problem attributes to case solution attributes.

However one should be careful to note that all of the learned similarity measure
are built on the assumption that similar data points have similar target values
(\(\delta_{s} \approx \delta_{e} \approx \delta_{p}\) in Figure
\ref{fig:problem-embedding-solution-space}). If this assumption does not hold,
learning the similarity measure might be much more difficult.

We have also presented a framework for how to analyze and group different types
of similarity measures. We have used this framework to analyze previous work and
highlight different strengths and weaknesses of the different types of
similarity measures. This also highlighted unexplored types of similarity
measures, such as Type 4 similarity measures. 

As a result we designed and evaluated a Type 3 similarity measure
\(t_{3,1}\) based on a classifier. The evaluations showed that using a classifier
as a basis for a similarity measure achieves comparable results to state of the
art methods, while using much less training evaluations to achieve that
performance.

We then combined strengths from Type 4 and Type 3 similarity measures
into a new Type 4 similarity measure, called Extended Siamese Neural
Networks (\(eSNN\)), which:
\begin{itemize}
\item Learns an embedding of the data points using \(G(\cdot)\) in the same way as
Type 3 similarity measures, but using shared weights in the same way as
SNNs to make the operation symmetrical.
\item Learns \(C(\hat{\vect{x}},\hat{\vect{y}})\), thus enabling extended performance in
relation to SNN and other Type 3 similarity measurements.
\item Restricts \(C(\hat{\vect{x}},\hat{\vect{y}})\) to make it invariant to input
ordering, and thus obtaining end to end symmetry through the similarity
measure.
\end{itemize}

Keeping \(eSNN\) symmetrical end-to-end enables the user of this similarity measure
to train on much smaller datasets than required by other types of similarity
measures. Type 3 measures based on SNNs also have this advantage, but our
results show that the ability to learn \(C(\hat{\vect{x}},\hat{\vect{y}})\) is
important for performance in many of the 14 datasets we tested. Our results
showed that \(eSNN\) outperformed state of the art methods on average over the 14
datasets by a large margin. We also demonstrated that \(eSNN\) achieved this
performance much faster given the same dataset than current state of the art. In
addition, the symmetry of \(eSNN\) enables it to train on datasets that are orders
of magnitude smaller. Our case-study of clustering embeddings produced from \(eSNN\)
show that the \(eSNN\) model can be used for semi-supervised clustering.

Finally we demonstrated that the training of this similarity measure scales to
large datasets like MNIST. Our main motivation for this work was to automate the
construction of similarity measures while keeping training time as low as
possible. We have shown that \(eSNN\) is a step towards this. Our evaluation shows
that it can learn similarity measures across a wide variety of datasets. We also
show that it scales well in comparison to similar methods and scales to datasets
of some size such as MNIST.

The applications for \(eSNN\) as a similarity measure are not only as a similarity
measure in a CBR system. It can also be used for semi-supervised clustering:
training \(eSNN\) on labeled data, then use the trained \(eSNN\) for clustering
unlabeled data. In much the same fashion it could be used for semi-supervised
clustering, using \(eSNN\) as a matching network in the same fashion as the distance
measure is applied in Vinyals et al. \cite{vinyals2016matching}.

In continuation of this work we would like to explore what is actually encoded
by learned similarity measures. This could be done by varying the different
features of a query data point \(\vect{q}\) in \(\mathbb{S}(\vect{x},\vect{q})\) and
discovering when that data point would change from one class to another (when
the class of the closest other data point changes) - this would form a
multi-dimensional boundary for each class. This boundary could be explored to
determine what the similarity measure actually encoded during the learning
phase.

Another interesting avenue of research would be to apply recurrent neural
networks to embed time series into embedding space (see Figure
\ref{fig:problem-embedding-solution-space}) to enable the similarity measure to
calculate similarity between time series which is currently a non-trivial
problem.

The architecture of similarity measures still require more investigation, e.g.
is the optimal embedding from \(G(\cdot)\) different from the softmax
classification vector used in normal supervised learning? If so it is worth
investigating why it is different.

\section{Acknowledgements}
\label{sec:orgb4c0436}
We would like to thank the EXPOSED project and NTNU Open AI Lab for the support
to do this work. Thanks also to Gunnar Senneset and Hans Vanhauwaert Bjelland
for their great support during our work.

\bibliography{index} 
\bibliographystyle{spbasic}
\end{document}